%% file: thbkg_arxiv_v2.tex
\documentclass[sigconf,nonacm]{acmart}

\usepackage{booktabs}      

\usepackage{amsmath, amssymb, bm}
\usepackage{multirow}
\usepackage{subcaption}
\usepackage{pifont}
\usepackage{array}
\usepackage{graphicx}

\providecommand{\colrule}{\midrule}
\providecommand{\botrule}{\bottomrule}
\providecommand{\articlesubtype}[1]{}

\newcommand{\cmark}{\ding{51}}
\newcommand{\xmark}{\ding{55}}
\newcommand{\hgt}{\textsc{HGT}}
\newcommand{\gatv}{\textsc{GATv2}}
\newcommand{\rgcn}{\textsc{R-GCN}}
\newcommand{\compgcn}{\textsc{CompGCN}}
\newcommand{\eahgt}{\textsc{EA-HGT}}

\begin{document}


\title{THBKG: A Temporal Biomedical Knowledge Graph for Decision-Aligned
Clinical Advancement Prediction}

\author{Pui Chung Siu}
\affiliation{%
  \institution{Queen Mary University of London}
  \department{School of Electronic Engineering and Computer Science}
  \city{London}
  \country{United Kingdom}}
\additionalaffiliation{%
  \institution{Recursion Pharmaceuticals Inc.}
  \city{Salt Lake City}
  \state{UT}
  \country{USA}}
\email{pui.siu@qmul.ac.uk}

\author{Claudia Cabrera}
\authornote{Corresponding authors.}
\affiliation{%
  \institution{Queen Mary University of London}
  \department{Clinical Pharmacology, William Harvey Research Institute,
              Barts and The London School of Medicine and Dentistry}
  \city{London}
  \country{United Kingdom}}
\email{c.cabrera@qmul.ac.uk}

\author{Mani Mudaliar}
\authornotemark[1]
\affiliation{%
  \institution{Recursion Pharmaceuticals Inc.}
  \city{Salt Lake City}
  \state{UT}
  \country{USA}}
\email{mani.mudaliar@recursion.com}

\author{Arkaitz Zubiaga}
\authornotemark[1]
\affiliation{%
  \institution{Queen Mary University of London}
  \department{School of Electronic Engineering and Computer Science}
  \city{London}
  \country{United Kingdom}}
\email{a.zubiaga@qmul.ac.uk}

\renewcommand{\shortauthors}{Siu et al.}

\begin{abstract}
Inadequate target--disease linkage accounts for 40--50\% of Phase~II efficacy
failures, so anticipating which programmes will advance would let sponsors back
the hypotheses most likely to reach patients. What a programme can be judged on
is the evidence that supported its linkage \emph{when it entered the clinic}. No
existing biomedical knowledge graph allows that evidence profile to be assembled
as of a past date. We present the Temporal Heterogeneous Biomedical Knowledge
Graph (THBKG), which describes and predicts therapeutic target--disease links
through time: 110,396 entities and 11.1M edges across nineteen relation types,
each edge carrying the year its evidence changed, so a pair's profile can be
recovered as it stood when its own decision fell due. On this graph we define a
decision-aligned benchmark that predicts, for a target--disease pair entering
Phase~II, whether it advances to Phase~III on evidence datable before that
decision. Graph propagation over the THBKG outranks every direct-evidence
reference scored under the same decision-aligned protocol, reaching a relative
success of 4.3--4.5 at the top ten pairs per therapeutic area. The gain
concentrates on the 72.8\% of pairs with no direct target--disease evidence at
their decision point, where a direct-edge model has nothing to read: the
encoders still rank five- to sixfold above chance, recovering the signal by
propagating over the intervening biology. Adapting a path-based explainer to the
decision-time subgraph decomposes each prediction into the evidence landscape
behind the hypothesis for explainable prediction.
We release the THBKG as a continually updated substrate for studying
therapeutic target hypotheses by retrospective validation.
\end{abstract}

\ccsdesc[500]{Applied computing~Biological networks}
\ccsdesc[500]{Information systems~Data mining}
\ccsdesc[500]{Information systems~Temporal data}

\keywords{Temporal Knowledge Graph, Biomedical Knowledge Graph,
Decision-Aligned Evaluation, Temporal Data Leakage, Clinical Trial
Prediction, Drug Target Identification}

\maketitle


\input{kdd_body/01_introduction}
\input{kdd_body/03_dataset_construction}
\input{kdd_body/04_task_and_models}

\input{kdd_body/05_experiments}

\input{kdd_body/06_discussion}
\input{kdd_body/07_conclusion}

\begin{acks}
This PhD project is jointly funded by the Biotechnology and Biological
Sciences Research Council (BBSRC) and Recursion Pharmaceuticals Inc. This
research utilised Queen Mary's Apocrita HPC facility, supported by
QMUL Research-IT.\footnote{\url{http://doi.org/10.5281/zenodo.438045}}
\end{acks}

\section*{Conflict of Interest}
P.C.S.\ and M.M.\ are affiliated with Recursion Pharmaceuticals Inc., a
drug-discovery company with a commercial interest in therapeutic target
identification. The THBKG is constructed entirely from publicly available
sources (Open Targets, ChEMBL, ClinicalTrials.gov) and is released under an
open licence; no proprietary Recursion data were used. The remaining authors
declare no competing interests.

\bibliographystyle{ACM-Reference-Format}
\bibliography{reference}

\appendix

\input{kdd_body/A_ethical_use_of_data}   
\input{kdd_body/B_datasheet}            
\input{kdd_body/C_graph_details}        
\input{kdd_body/D_models}               
\input{kdd_body/E_protocol}             
\input{kdd_body/F_results_figures}      
\clearpage
\input{kdd_body/G_case_studies}         

\end{document}

%% file: kdd_body/01_introduction.tex
%
%

\section{Introduction}

Late-stage clinical failure is the dominant cost in drug development, driven
principally by targets not causally linked to their intended disease.
Inflation-adjusted cost per approved drug
has doubled roughly every nine years since 1950, a trend known as Eroom's Law
\cite{Scannell2012DiagnosingEfficiency}, and inadequate target--disease
linkage accounts for 40--50\% of Phase~II efficacy failures
\cite{Sun2022WhyIt, Arrowsmith2013Trial2011-2012}. Prioritsing targets that
drive a disease from those merely associated with it is therefore the central
problem of therapeutic target identification, and the strongest empirical
handle on it is genetic: drugs whose target--disease hypotheses carry human
genetic support succeed in the clinic at roughly twice the rate of those
without \cite{Nelson2015, Hingorani2019, Schmidt2020, Trajanoska2023,
Minikel2024}.

The evidence bearing on a target--disease hypothesis is rarely attached to the
hypothesis itself, because disease seldom traces to a single gene but to
perturbations propagating through the cellular network linking genes, pathways
and phenotypes \cite{Barabasi2011NetworkMedicine}, which reach a target through
intermediate biology rather than through a direct association. Biomedical
knowledge graphs integrate genetic, molecular, pathway and clinical
sources into one structure, making the full support behind a hypothesis legible
and reachable along multi-hop paths
\cite{Chandak2023PrimeKG, Narganes-Carlon2024GATher:Links}.

Evidence does not arrive all at once but accrues over decades, each item
entering the record when the observation behind it is first reported, so the
support a target--disease link carries grows as genetic, literature and clinical
sources accumulate at different rates \cite{Falaguera2025TemporalDiscovery}. A decision taken in 2016
was taken on the evidence that existed in 2016; scoring it on a graph built in
2026 credits the model with a decade of evidence the decision-maker never had,
much of which accrued \emph{because} the decision was taken. 
To our knowledge, no biomedical knowledge graph enables the evidence subgraph to
be recovered as it stood at a past event-time cutoff
(Table~\ref{tab:resource_comparison}). Static graphs
\cite{Chandak2023PrimeKG, Himmelstein2017Hetionet, Walsh2020BioKG, Breit2020OpenBioLink} carry no edge
timestamps, so they are frozen at whenever their source databases were pulled; GATher
\cite{Narganes-Carlon2024GATher:Links} does consider temporality for retrospective validation, by freezing the
graph at a single global cutoff that cannot afterwards be re-cut; MedKGent
\cite{Zhang2025MedKGent} does date its edges by the publication of the abstract publication date but temporality is eventually dissolved to latest status. Falaguera
et al.\ \cite{Falaguera2025TemporalDiscovery} timestamp Open Targets evidence and
recover yearly association trajectories, but their analysis is post hoc and confined to
direct target--disease evidence.

This absence has shaped how the field evaluates downstream tasks such as novel target identification and drug repurposing. Biomedical knowledge graph
link prediction is assessed predominantly under random, transductive edge
holdout, in which a fact is hidden and the model asked to recover it while the
surrounding structure is held fixed; the alternative hides entities
rather than edges \cite{Bang2025Inductive}. Neither reconstructs the
evidentiary state of the world at the moment the finding was actionable. By contrast, the
general-domain temporal knowledge graph literature evaluates under
chronological splitting \cite{Jin2020RENET}, and its
interpolation/extrapolation distinction
\cite{GarciaDuran2018, Cai2024TKGSurvey} exists because forecasting
requires the future to be withheld; tho no resources are available for the biomedical domain. Biomedical graphs run interpolation-style
random splits while making extrapolation-style claims about discovery, an
instance of what Kapoor and Narayanan classify as temporal leakage: a test set
not drawn from the distribution about which the scientific claim is made
\cite{Kapoor2023Leakage, Hu2020OGB}.

\begin{table}[t]
\caption{Capabilities of biomedical and temporal knowledge-graph resources.
\emph{Multi-hop}: heterogeneous message passing across entity types, so
evidence not attached to a pair can still reach it. \emph{Event-time}: edges
carry the date the underlying observation was first reported. \emph{As-of}:
the graph can be reconstructed at an arbitrary cutoff, which the query rather
than the resource chooses. Every prior resource offers at most two of the
three. Notes: (a) roll-up along the disease ontology only, not message passing
across entity types; (b) time is controlled, but by a single global freeze
(2018-01-01) that cannot be re-cut; (c) dated by the publication of the source
abstract, but extracted from literature only; (d)
reconstructs yearly association trajectories \emph{post hoc}, as an analysis of
evidence accrual rather than a substrate for prediction.}
\label{tab:resource_comparison}
\centering
\small
\begin{tabular}{@{}lccc@{}}
\toprule
\textbf{Resource} & \textbf{Multi-} & \textbf{Event-} & \textbf{As-} \\
                  & \textbf{hop}    & \textbf{time}   & \textbf{of}  \\
\midrule
PrimeKG \cite{Chandak2023PrimeKG}                & \cmark & \xmark & \xmark \\
Open Targets \cite{Ochoa2021OpenPrioritisation}  & \xmark$^{a}$ & \xmark & \xmark \\
GATher \cite{Narganes-Carlon2024GATher:Links}    & \cmark & \xmark & \xmark$^{b}$ \\
MedKGent \cite{Zhang2025MedKGent}                & \xmark$^{c}$ & \cmark & \xmark \\
OT time-series \cite{Falaguera2025TemporalDiscovery} & \xmark$^{a}$ & \cmark & \xmark$^{d}$ \\
Czech et al.\ \cite{Czech2024CLINICALFORECASTING}    & \xmark & \cmark & \cmark \\
\midrule
\textbf{THBKG (ours)}                            & \cmark & \cmark & \cmark \\
\bottomrule
\end{tabular}
\end{table}

The task we address is clinical advancement prediction \cite{Czech2024CLINICALFORECASTING}: for a target--disease pair
first entering Phase~II, whether it advances to Phase~III, scoring it only on
evidence that existed at that decision. Their ridge regression over 33
engineered evidence features identifies the top 2\% of pairs at 4--5$\times$ the
base advancement rate using biomedical evidence alone, and they establish the
protocol that makes such a result meaningful: a calendar split separating train
and test pairs in time, and a decision-aligned split further restricting each
pair to evidence predating its own Phase~II entry. Under that masking the great majority of pairs carry
no evidence linking them at all, so their model relies on target-level and
disease-level priors, and largely answers whether a \emph{target} has clinical
precedent rather than whether \emph{this} target--disease hypothesis is
supported.

We present the Temporal Heterogeneous Biomedical Knowledge Graph
(THBKG), which assembles timestamped genetic, literature, pathway, interaction
and clinical-trial evidence into a heterogeneous graph whose every temporal
edge carries the year its evidence changed (1995--2025). Masking the
edges that postdate a pair's Phase~II entry recovers the evidence landscape as
it stood at that pair's own decision point, and the graph's multi-hop structure
survives the decision-aligned masking. To
characterise what that structure supports, we evaluate the main families of
heterogeneous encoder on it, together with an edge-aware variant that extends
attention \cite{Hu2020HeterogeneousTransformer} to weight each association by the
strength and the recency of the evidence behind it.

This work makes three contributions.

\begin{enumerate}
\item The \textbf{THBKG} knowledge graph and its construction pipeline. Every edge is dated by
the event time of the observation behind it, so the graph answers as-of queries
at any cutoff. Because the pipeline indexes evidence by event time rather than
ingesting a snapshot, evidence reported later extends the index rather than
invalidating it, and the substrate can be extended as the record accrues. This
makes retrospective validation available to any biomedical hypothesis carrying a
decision date, novel target identification and drug repurposing among them.

\item \textbf{A decision-aligned advancement benchmark.} We promote that task to a
benchmark: each target--disease pair is scored only from evidence datable strictly
before its own clinical transition year, and we release fixed splits, a
therapeutic-area--aware relative-success protocol, and a characterised suite of
reference conditions spanning tabular models, four heterogeneous encoder families,
and an edge-aware variant on two backbones. The graph encoders enrich
the top of the ranking over the tabular baselines, but the gain is a property of
relation-aware propagation rather than of any one architecture, and it is not
uniform: it concentrates on pairs carrying no direct target--disease evidence,
whose support is reachable only through the intervening biology, and is not
resolvable on targets reaching Phase~II for the first time, the stratum a
discovery programme would most want and the one the record is thinnest on
(Section~\ref{sec:main_results}).

\item \textbf{Decomposable predictions.} A decision-point subgraph decomposes
into the individual dated observations that produced the score and the multi-hop
route they travelled, so a ranking can be audited against the evidence that
produced it and the year each item entered the record
(Appendix~\ref{app:explainability}).

\end{enumerate}


%% file: kdd_body/03_dataset_construction.tex
%
%

\section{The THBKG Knowledge Graph}
\label{sec:thbkg}

\subsection{THBKG Construction}
\label{sec:thbkg_construction}

\begin{figure}[t]
\centering
\begin{subfigure}[b]{\columnwidth}
\centering
\includegraphics[width=\columnwidth]{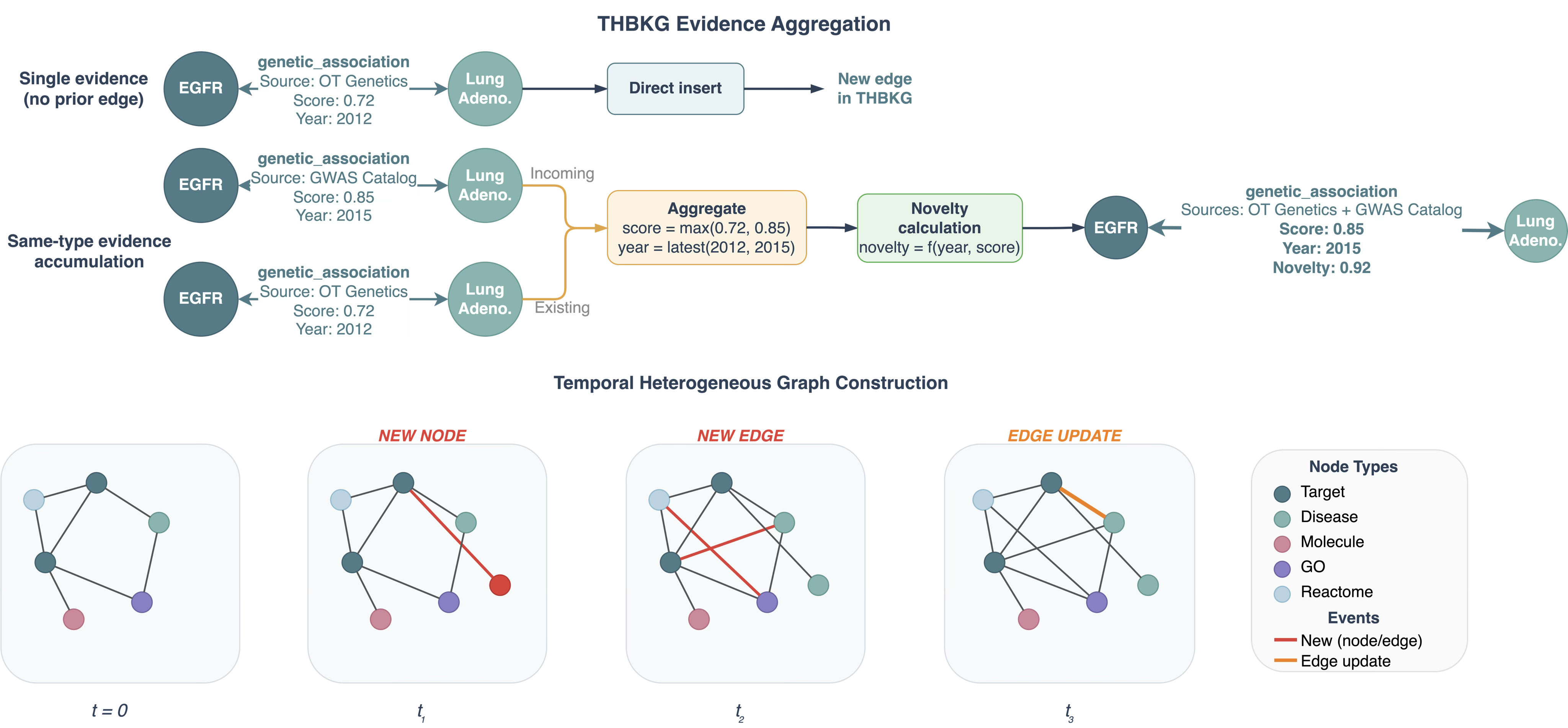}
\caption{}
\label{fig:thbkg_construction}
\end{subfigure}

\vspace{0.8em}

\begin{subfigure}[b]{\columnwidth}
\centering
\includegraphics[width=\columnwidth]{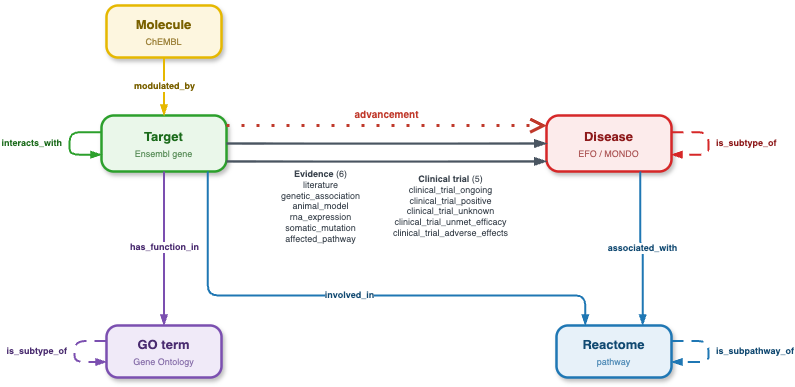}
\caption{}
\label{fig:thbkg_schema}
\end{subfigure}
\caption{The THBKG. \textbf{(a)} Construction pipeline: dated evidence records
are scored by harmonic aggregation and assembled into a heterogeneous graph.
\textbf{(b)} Schema: five node types (target, disease, molecule, Reactome
pathway, GO term) connected by 16 temporal edge types, subject to per-instance
masking, and 3 static ontology edge types, present regardless of cutoff.}
\label{fig:thbkg}
\end{figure}

The THBKG comprises five node types and nineteen context relations, sixteen of
them dated and three carrying the definitional ontology structure
(Figure~\ref{fig:thbkg}, Appendix Table~\ref{tab:edge_types_full}), assembled from four
sources ingested and dated independently. Its schema adapts that of GATher
\cite{Narganes-Carlon2024GATher:Links}.

We retrieve target--disease evidence from Open Targets release 26.03
\cite{Ochoa2021OpenPrioritisation}, which aggregates literature, genetic
association, animal model, RNA expression, somatic mutation and affected-pathway
evidence; the contributing datasources for each are given in Appendix
Table~\ref{tab:edge_types_full}. Each record is dated following the method of
Falaguera et al.\ \cite{Falaguera2025TemporalDiscovery}.

The remaining relations connect targets to one another and to pathways, functions
and compounds. Protein--protein interactions are
parsed from the IntAct MITAB
release \cite{Orchard2014IntAct}, restricted to protein--protein pairs and dated
by the earliest publication reporting each interaction, resolved from the PubMed
identifiers on the record. Functional annotation is taken from the Gene Ontology
\cite{Ashburner2000GeneOntology} and pathway membership from Reactome
\cite{Gillespie2022Reactome}. Annotations carrying no
resolvable reference have no event time and are not admitted to the graph. The disease, Gene Ontology and Reactome
hierarchies are ingested as static structure (Section~\ref{sec:thbkg_features}).

Clinical-trial records are drawn from the \texttt{clinical\_%
precedence} evidence of Open Targets, which carries the ClinicalTrials.gov registration behind
each trial; records without a registry identifier are dropped. Each trial is split into an ongoing edge at its
start date and an outcome edge at its terminal date, so that an outcome cannot
enter the graph before it was reported; the date-resolution procedure is given in
Appendix~\ref{sec:supp_ct_dates}. The same records supply the
\texttt{modulated\_by} relation connecting a drug to the target it acts on, so
this relation reflects clinical development rather than binding affinity. These
relations are retained as context, since prior clinical precedent was available
to a decision-maker. The supervision label (Section~\ref{sec:advancement_labels})
is stored as a \texttt{target $\rightarrow$ disease} edge but excluded from
message passing.

The pipeline requires only that a record carries a date, an entity pair and a
score, so sources may be substituted or added, and evidence reported later
extends the index rather than invalidating it.

\paragraph{Temporal evidence aggregation}
\label{sec:ch4-scoring}

The cumulative score $s_{ij}$ follows the Open Targets harmonic-sum convention
\cite{Ochoa2021OpenPrioritisation}. For evidence items $s_1\geq\cdots\geq s_n$
with year $\leq Y$,
\begin{equation}
    H = \frac{1}{\zeta(2)} \sum_{i=1}^{\min(n,\,50)} \frac{s_i}{i^2},
    \qquad \zeta(2) = \frac{\pi^2}{6} \approx 1.644,
    \label{eq:harmonic}
\end{equation}
where the $i^{-2}$ weighting lets the strongest items dominate and $\zeta(2)$
normalises $H\in[0,1]$. Clinical-trial evidence is an ordinal phase ceiling
rather than accumulating support, and is aggregated by maximum instead.

The novelty score $n_{ij}$ \cite{Falaguera2025TemporalDiscovery} is read from
the trajectory itself. With $\Delta_p=\max(s_p-s_{p-1},0)$ the positive
year-on-year jump at year $p$,
\begin{equation}
    n_{t} = \max_{\,t - W \leq p \leq t}\;
        \frac{\Delta_p}{1 + \exp\!\bigl(k\,((t - p) - m)\bigr)},
        \qquad k = 2,\; m = 3,\; W = 5,
    \label{eq:novelty}
\end{equation}
so novelty peaks just after evidence arrives and decays as it ages. The two
features encode complementary quantities: $s_{ij}$ how much support exists,
$n_{ij}$ when it arrived.

\paragraph{Event compression}

To avoid storing unchanged edges across years, an edge is written only in years where the
cumulative score differs from the prior year, so every stored edge corresponds to
a genuine update in the evidence record. The encoding is lossless: for any cutoff
year $y$, a triplet's value is the most recent stored edge with $t_{ij} < y$.

\paragraph{The as-of query}

Because every edge carries an event time, the graph is queryable in time: for a
cutoff year $y$, the \emph{as-of query} $\mathcal{G}[y]$ returns the subgraph
induced by the edges whose evidence predates $y$,
\begin{equation}
    \mathcal{G}[y] \;=\; \bigl(\mathcal{V},\;
    \{\, e_{ij} \in \mathcal{E} \;:\; t_{ij} < y \,\}\bigr),
    \label{eq:asof}
\end{equation}
with static ontology edges carrying a sentinel timestamp and never masked.
Since $y$ is a free parameter, the regime is set by the task rather than by the
resource: a study with one global horizon fixes a single $y$, whereas a study
whose instances were decided on different dates sets $y$ per instance. The
regime this task requires is given in Section~\ref{sec:formulation}.

\paragraph{Timestamp assignment}

Static structural edges, the disease, Gene Ontology and Reactome hierarchies,
carry a sentinel timestamp and are present regardless of cutoff, being
definitional rather than evidential. All other edges are dated by the source that
produced them and carry $\mathbf{f}_{ij} = [s_{ij}, n_{ij}]$ as their attribute
vector.

\paragraph{Node feature engineering}
\label{sec:thbkg_features}

Each node type is initialised with a domain-specific feature vector,
following \citet{Narganes-Carlon2024GATher:Links}: targets ($d=56$) from
expression specificity and genetic-constraint scores, diseases ($d=256$)
from biomedical language-model embeddings of their EFO names and
descriptions, GO terms and Reactome pathways ($d=64$ each) from
graph-auto-encoder embeddings of their ontology hierarchies, and molecules
($d=1{,}024$) from radius-2 Morgan fingerprints. The per-feature
construction is detailed in Appendix Section~\ref{sec:supp_features}.


%% file: kdd_body/04_task_and_models.tex
%
%

\section{Task and Models}
\label{sec:task}

\subsection{Clinical Advancement Labels}
\label{sec:advancement_labels}

We derive binary advancement labels following the per-instance,
decision-aligned protocol of Czech et al.\
\cite{Czech2024CLINICALFORECASTING}. From ClinicalTrials.gov we take all
interventional trials with a registered Phase~II arm, map each
intervention to an Ensembl gene target through ChEMBL
\cite{Gaulton2017The2017} and each condition to an EFO disease term, and
retain only pairs with a unique target and disease; multi-mechanism
combination trials are excluded. A pair is labelled positive if a
Phase~III trial for the same target--disease combination is registered
within three years of its Phase~II entry, a window matched to the
typical Phase~II-to-Phase~III progression time, around which roughly half
of Phase~II trials complete \cite{Wong2019}, and negative otherwise.
Pairs with ambiguous status (open observation windows, or Phase~II
terminations for non-efficacy reasons such as safety or commercial
decisions) are excluded to limit label noise. Advancement is a
low-base-rate event, consistent with reported Phase~II-to-Phase~III
attrition \cite{Wong2019}: 22.76\% of training pairs advance against 9.29\% of
evaluation pairs, since a pair entering Phase~II late in the evaluation window
has less of its three-year observation period elapsed and falls to the negative
class (Appendix Table~\ref{tab:traineval_ta}).

 Advancement is defined over the target--disease therapeutic hypothesis rather than the
molecule, a pair advances when \emph{any} drug testing the hypothesis progresses,
so the label is robust to molecule-level failure and tracks the field's sustained
willingness to keep testing the link \cite{Hwang2016LateStageFailure}.


\subsection{Problem Formulation}
\label{sec:formulation}

Let $\mathcal{G} = (\mathcal{V}, \mathcal{E}, \mathcal{A}, \mathcal{R})$ denote
a temporal heterogeneous knowledge graph with node- and edge-type maps
$\tau(v) : \mathcal{V} \rightarrow \mathcal{A}$ and
$\phi(e) : \mathcal{E} \rightarrow \mathcal{R}$, so that each edge $e_{ij}$
carries the meta-relation
$\langle \tau(v_i),\, \phi(e),\, \tau(v_j) \rangle$. Each edge additionally
carries an \emph{event time} $t_{ij}$, the year the underlying observation
was first reported, and the feature vector
$\mathbf{f}_{ij} = [s_{ij},\, n_{ij}]^\top$ of
Section~\ref{sec:thbkg_construction} (Equations~\eqref{eq:harmonic}
and~\eqref{eq:novelty}).

The as-of query $\mathcal{G}[y]$ (Equation~\eqref{eq:asof}) implements the
decision-aligned protocol of Czech et al.\ \cite{Czech2024CLINICALFORECASTING}
over the graph. Each pair $(v_t, v_d)$ is dated by its first Phase~II entry,
$y_{td}$, and appears once. Scoring it on $\mathcal{G}[y_{td}]$ restricts it to
the evidence that existed when the hypothesis first reached the clinic. Applying it to a heterogeneous
multi-hop graph requires the graph itself to carry event times on every edge.

\subsection{Baseline Models}
\label{sec:baselines}

We compare against four heterogeneous encoders spanning the main families
of relation-aware message passing: \hgt{}
\cite{Hu2020HeterogeneousTransformer}, which parameterises attention with
node- and edge-type-specific projections indexed by the meta-relation
triple; \rgcn{} \cite{Schlichtkrull2018ModelingRelational}; \compgcn{}
\cite{Vashishth2020CompositionBased}; and \gatv{}
\cite{Brody2021HowNetworks}. All four condition attention and messages on
relation \emph{type} but are blind to the continuous edge features
$\mathbf{f}_{ij}$: two edges of the same relation type receive identical
weight regardless of their evidential score or novelty. Beyond the graph encoders we compare against the
ridge regression of Czech et al.\ \cite{Czech2024CLINICALFORECASTING}
(RDG), the strongest direct-edge baseline, and the Open Targets global
association score (OTS). Both are regenerated on this release and evaluation
split rather than transcribed from their published values
(Appendix~\ref{app:protocol}), so every condition is scored on the same pairs. Architectures and hyperparameters for all
conditions are given in Appendix Table~\ref{tab:hyperparams}.

\paragraph{Edge-aware variant}
The four encoders are blind to $\mathbf{f}_{ij}$, so testing whether the
evidential features carry signal requires a variant that reads them. The
Edge-Aware Heterogeneous Graph Transformer (\eahgt{}) modulates \hgt{}'s
attention by the score and novelty on each surviving edge, adding two parameters
per relation type (Appendix~\ref{app:eahgt}); the same modulation applied to
\gatv{} gives a $2 \times 3$ ablation over backbone and feature subset. These are
instruments for that ablation, not a proposed state of the art.

\subsection{Evaluation Protocol}
\label{sec:training_evaluation}

All conditions share one training and ensembling recipe, one temporal train/test
split, and one therapeutic-area selection, specified in
Appendix~\ref{app:protocol}. The evidence strata and the ranking metric are
defined here.

\paragraph{Evidence stratification}
\label{sec:strata}

To localise where the graph gains over a direct-feature baseline, we stratify the
evaluation set along two independent axes, each computed on evidence dated before
the pair's Phase~II entry year, and report the pooled set ($n = 7{,}193$)
alongside them.

The \emph{evidence} axis isolates the evidence-sparse regime: \emph{no prior
evidence} pairs carry no \texttt{genetic\_association}, \texttt{somatic\_mutation},
\texttt{affected\_pathway}, \texttt{animal\_model}, \texttt{rna\_expression} or
\texttt{literature} edge to their disease at their decision point ($n = 5{,}239$,
72.8\%). The \emph{target-history} axis splits pairs by whether their target had
previously entered Phase~II for any disease ($n = 7{,}023$) or had not
($n = 170$).

Every evaluation pair entered Phase~II by selection, so a clinical-trial edge to
its own disease is near-universal. It cannot discriminate advancers from
non-advancers and is excluded from both axes: a pair with \emph{no prior
evidence} may still carry the trial edge that put it in the evaluation set.

\paragraph{Metrics}
\label{sec:metrics}

We adopt the evaluation protocol of Czech et al.\
\cite{Czech2024CLINICALFORECASTING} in full. Relative success at $N$ (RS@$N$) is
the advancement rate among the top-$N$ ranked pairs over that among the rest,
\begin{equation}
    \text{RS@}N =
    \frac{P(\text{advancement} \mid \text{rank} \leq N)}
         {P(\text{advancement} \mid \text{rank} > N)},
    \label{eq:rsn}
\end{equation}
so RS@$N = 4$ means the top-$N$ pairs advance at four times the rate of the
remainder. The metric has precedent in genetic-support benchmarking
\cite{Nelson2015, King2019, Minikel2024}.

The primary metric is the \emph{therapeutic-area mean} of RS@$N$:
Equation~\eqref{eq:rsn} is evaluated within each of the 13 retained therapeutic
areas and the per-area values equally weighted, so that areas contribute
comparably regardless of size. We report it at
$N \in \{10, 20, 30, 40, 50, 100\}$ and at the top-1\% and top-2\% cutoffs. An
equally-weighted mean over few, high-variance areas is readily dominated by one
or two of them \cite{Agarwal2021StatisticalPrecipice}, so we report the per-area
median and the number of areas beating the baseline alongside it, and contrast
the three aggregations in the ablation (Section~\ref{sec:ablation}).

\textbf{AUROC} and \textbf{Average Precision (AP)} are reported as
secondary metrics.

%% file: kdd_body/05_experiments.tex
%
%

\section{Experiments}

\subsection{The Temporal Heterogeneous Biomedical Knowledge Graph}
\label{sec:graph_characterisation}

The THBKG integrates timestamped evidence into 16 temporal and 3 static context relation types, spanning
target--disease associations accumulated over two decades. Its static edge
composition is markedly heterogeneous (Appendix
Figure~\ref{fig:thbkg_sunburst}): literature and genetic association account
for the plurality of the target--disease evidence, while clinical and pathway
evidence occupy much smaller fractions, and protein--protein interaction,
functional annotation, pathway membership, and drug modulation edges supply
the structural backbone for multi-hop propagation.

Table~\ref{tab:asof_growth} makes the as-of query concrete by applying it at
five cutoffs. The graph grows by roughly an order of magnitude between 2005 and
2025, but the composition shifts as it grows: genetic and causal evidence
expands a hundredfold over the period while the experimental and functional
backbone little more than doubles, so an early cut is not a scaled-down copy of
the present-day graph but a differently constituted one. A model scored at 2010
therefore sees a graph in which literature co-mention supplies the majority of
target--disease evidence and genetic support is comparatively scarce.

\begin{figure}[t]
\centering
\includegraphics[width=\columnwidth]{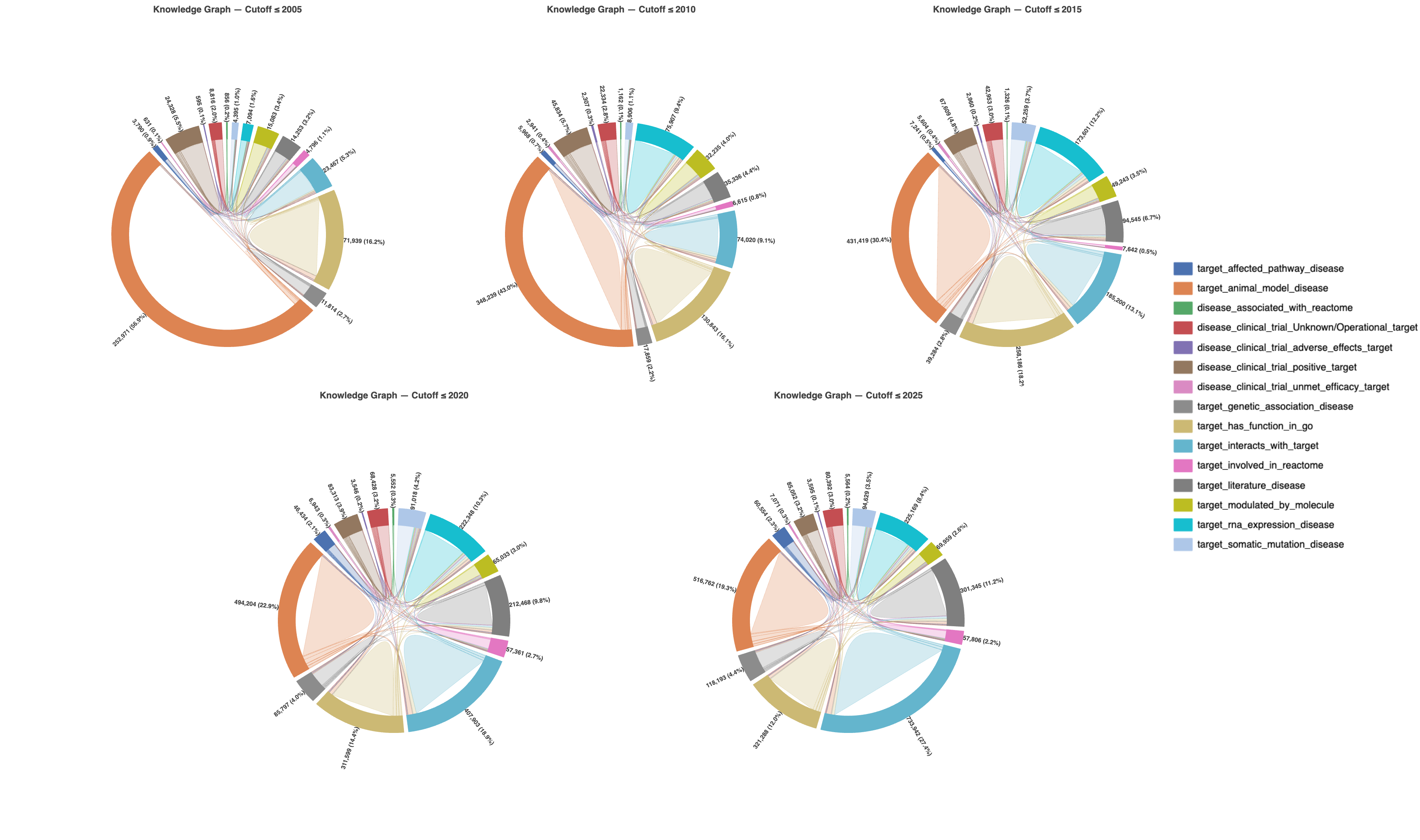}
\caption{Temporal edge distribution of the THBKG. Chord diagrams at five
cumulative temporal cutoffs ($\leq$2005, $\leq$2010, $\leq$2015, $\leq$2020,
$\leq$2025). Each node represents a directed edge-type triple; arc size is
proportional to edge count at that cutoff, and ribbons connect triples that
co-occur on shared entities, weighted by their minimum joint count.}
\label{fig:thbkg_chord}
\end{figure}

Viewed over time, the graph densifies in a structured, datatype-specific way.
The cumulative chord diagrams (Figure~\ref{fig:thbkg_chord}, five cutoffs from
$\leq$2005 to $\leq$2025) show animal-model and early literature evidence
dominating the earliest snapshot, with somatic-mutation, RNA-expression and
protein--protein interaction edges accruing later. Each edge type enters the
record once the technology that produces it exists, and lands on the targets and
diseases the field was studying at the time
\cite{Falaguera2025TemporalDiscovery}; the composition of the graph at any year
is thus a record of what biomedical research had discovered, and where it had
been looking. Evidence accrues continuously as that research is done, so a cutoff
does not select a smaller version of the present-day graph but the state of an
ongoing process at a particular moment.

\begin{table}[htbp]
\centering
\caption{Effect of the as-of query on graph size. Each column re-cuts the graph
at a decision date $\tau$, keeping only evidence timestamped $\le\tau$; the
2025 cut is the full graph. Edge counts in millions, node counts in thousands.
Dated evidence only: the static ontology backbone ($125{,}110$ edges, never
masked) and the advancement labels ($28{,}795$, excluded from message passing)
account for the difference from the $11.1$M total of
Section~\ref{sec:thbkg_construction}. Totals are computed before rounding.}
\label{tab:asof_growth}
\small
\setlength{\tabcolsep}{4.5pt}
\begin{tabular}{@{}lrrrrr@{}}
\toprule
& \multicolumn{5}{c@{}}{As-of cut $\tau$} \\
\cmidrule(l){2-6}
& 2005 & 2010 & 2015 & 2020 & 2025 \\
\midrule
\multicolumn{6}{@{}l}{\emph{Evidence edges} (M)} \\
\quad Genetic \& causal        & 0.02 & 0.03 & 0.09 & 0.35 & 1.76 \\
\quad Literature co-mention    & 0.77 & 1.38 & 2.52 & 4.35 & 7.27 \\
\quad Experimental \& funct.   & 0.38 & 0.56 & 0.74 & 0.82 & 0.88 \\
\quad Clinical                 & 0.03 & 0.07 & 0.13 & 0.20 & 0.25 \\
\quad Pathway \& molecular     & 0.06 & 0.13 & 0.25 & 0.47 & 0.80 \\
\textbf{Total}                 & \textbf{1.26} & \textbf{2.17} & \textbf{3.73} & \textbf{6.19} & \textbf{10.95} \\
\midrule
\multicolumn{6}{@{}l}{\emph{Nodes reached}$^{\dagger}$ (K)} \\
\quad target                   & 13.5 & 17.0 & 18.7 & 19.4 & 19.6 \\
\quad disease                  & 13.9 & 14.8 & 16.3 & 18.5 & 26.3 \\
\quad molecule                 &  1.1 &  1.9 &  2.3 &  2.6 &  2.6 \\
\bottomrule
\end{tabular}

\vspace{2pt}
{\footnotesize $^{\dagger}$Nodes incident to at least one dated edge $\le\tau$.
GO ($38.5$K) and reactome ($2.6$K) nodes attach through the static backbone and
are time-invariant. A further $20.8$K disease terms attach by ontology alone,
giving $47.0$K in the released graph.}
\end{table}

\subsection{Multi-Hop Propagation on the THBKG Improves Top-of-Ranking Enrichment over a Direct-Edge Baseline}
\label{sec:main_results}

We evaluate the graph encoders against the ridge regression baseline of Czech et al.\
\cite{Czech2024CLINICALFORECASTING} (RDG) and the Open Targets global association
score (OTS) on the held-out evaluation set of $N = 7{,}193$ Phase~II
target--disease pairs (9.29\% positive).

\paragraph{Relative success at top-$N$}

\begin{figure}[t]
\centering
\begin{subfigure}[t]{0.62\columnwidth}
\centering
\includegraphics[width=\textwidth]{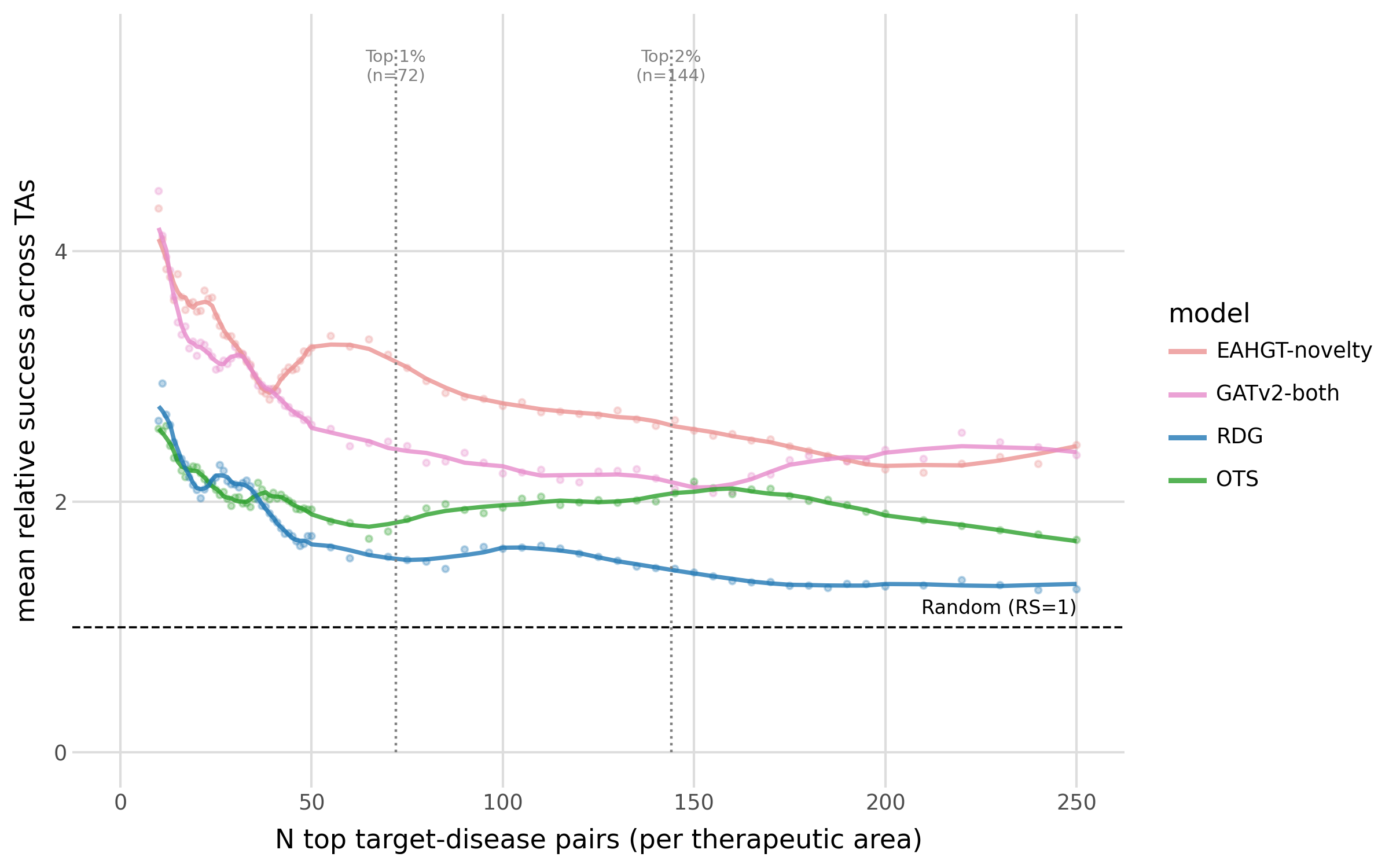}
\caption{}
\label{fig:rs_by_limit}
\end{subfigure}%
\begin{subfigure}[t]{0.38\columnwidth}
\centering
\includegraphics[width=\textwidth]{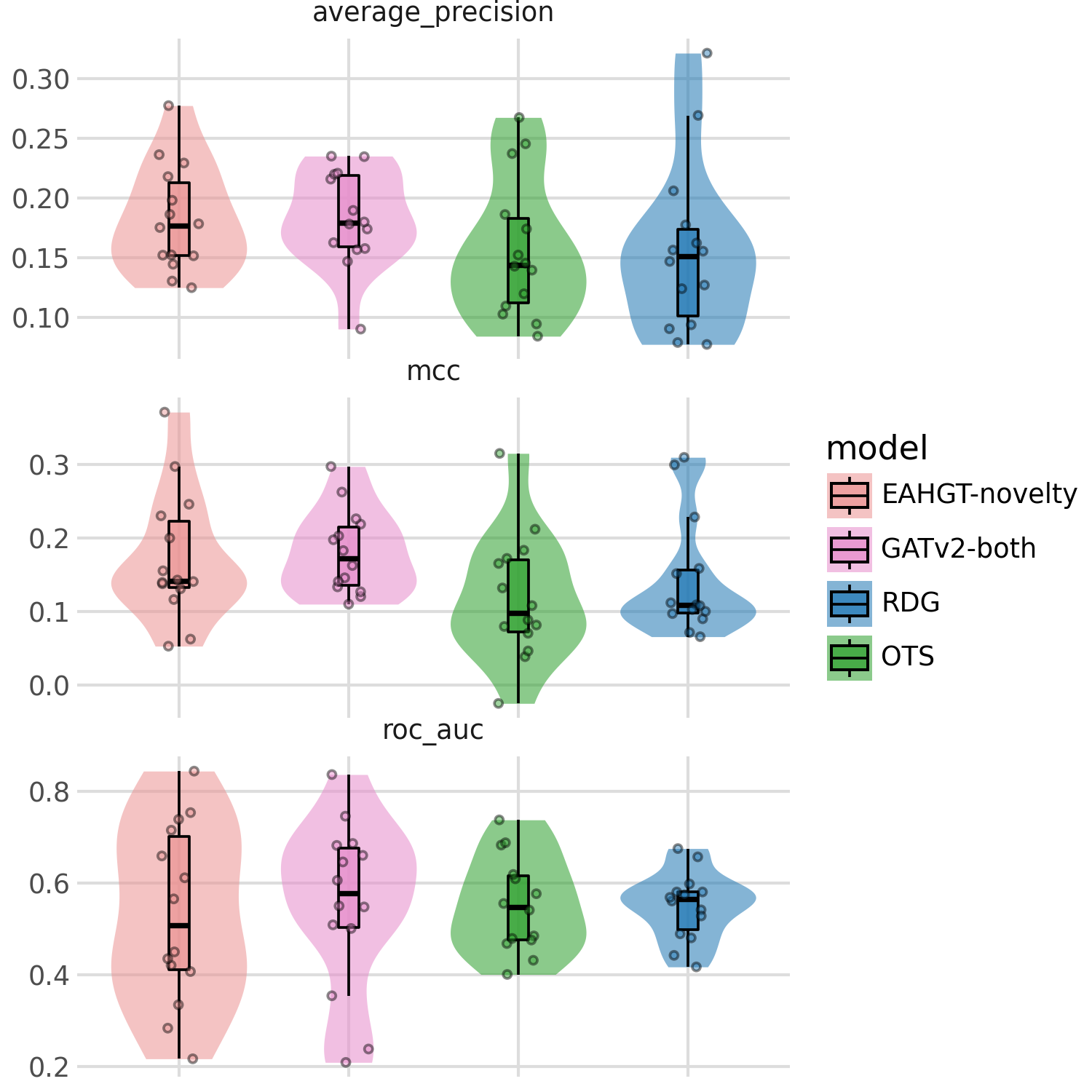}
\caption{}
\label{fig:classification_by_ta}
\end{subfigure}
\caption{Ranking and classification performance of the graph encoders
against the RDG and OTS reference baselines, under decision-aligned
evaluation (Open Targets 26.03, evaluation window w3).
\textbf{(a)} Therapeutic-area mean relative success as a function of
top-$N$ cutoff, averaged across 13 therapeutic areas; dotted lines mark the
top 1\% and top 2\% of test pairs, the dashed line marks random ranking
(RS $=1$). The graph encoders lead the tabular baselines across the
actionable range; which encoder leads depends on the cutoff.
\textbf{(b)} Classification metrics (average precision, MCC, and ROC-AUC) by
therapeutic area; each violin summarises the per-area distribution with an
inner box for the median and IQR, and points are per-area values. Both panels
show the best condition per encoder family against the tabular references; the
full twelve-condition set is given in Appendix Figure~\ref{fig:agg_divergence}
and Table~\ref{tab:ablation}.}
\label{fig:rs_and_classification}
\end{figure}

At the top of the ranking the graph encoders lead RDG, the strongest published
model for this task, and OTS (Table~\ref{tab:ablation},
Figure~\ref{fig:rs_by_limit}). At $N = 10$ the strongest reach RS $\approx$
4.3--4.5 against 2.65 for RDG and 2.58 for OTS; at $N = 50$, RS $\approx$ 3.2
against $\approx$ 1.9; and at $N = 100$, RS $\approx$ 2.8 against $\approx$ 1.8.
The lead holds across the actionable range and for every encoder at $N = 50$,
with two exceptions at the extremes: \compgcn{} falls below RDG at $N = 10$
(2.00 against 2.65) and \rgcn{} matches it at $N = 100$ (1.62). Which encoder
leads depends on the cutoff.

\paragraph{Global classification metrics}

Global discrimination separates the models far less, pooled over all evaluation pairs the whole suite sits in a narrow
band above chance (AUROC 0.52--0.62 against 0.57 for RDG and 0.55 for OTS), with
average precision uniformly low (0.11--0.18 against a 9.3\% base rate). The
$\approx$ 0.1 AUROC spread is small next to the RS@10 spread and does not
separate the encoders from the tabular references. The value the encoders add is
concentrated at the top of the ranking, where prioritisation decisions are made,
and is only weakly visible in threshold-free global metrics, consistent with
Czech et al.'s observation \cite{Czech2024CLINICALFORECASTING}. The per-therapeutic-area
breakdown is given in Appendix~\ref{app:aggregate}.

\begin{figure}[t]
\centering
\includegraphics[width=\columnwidth]{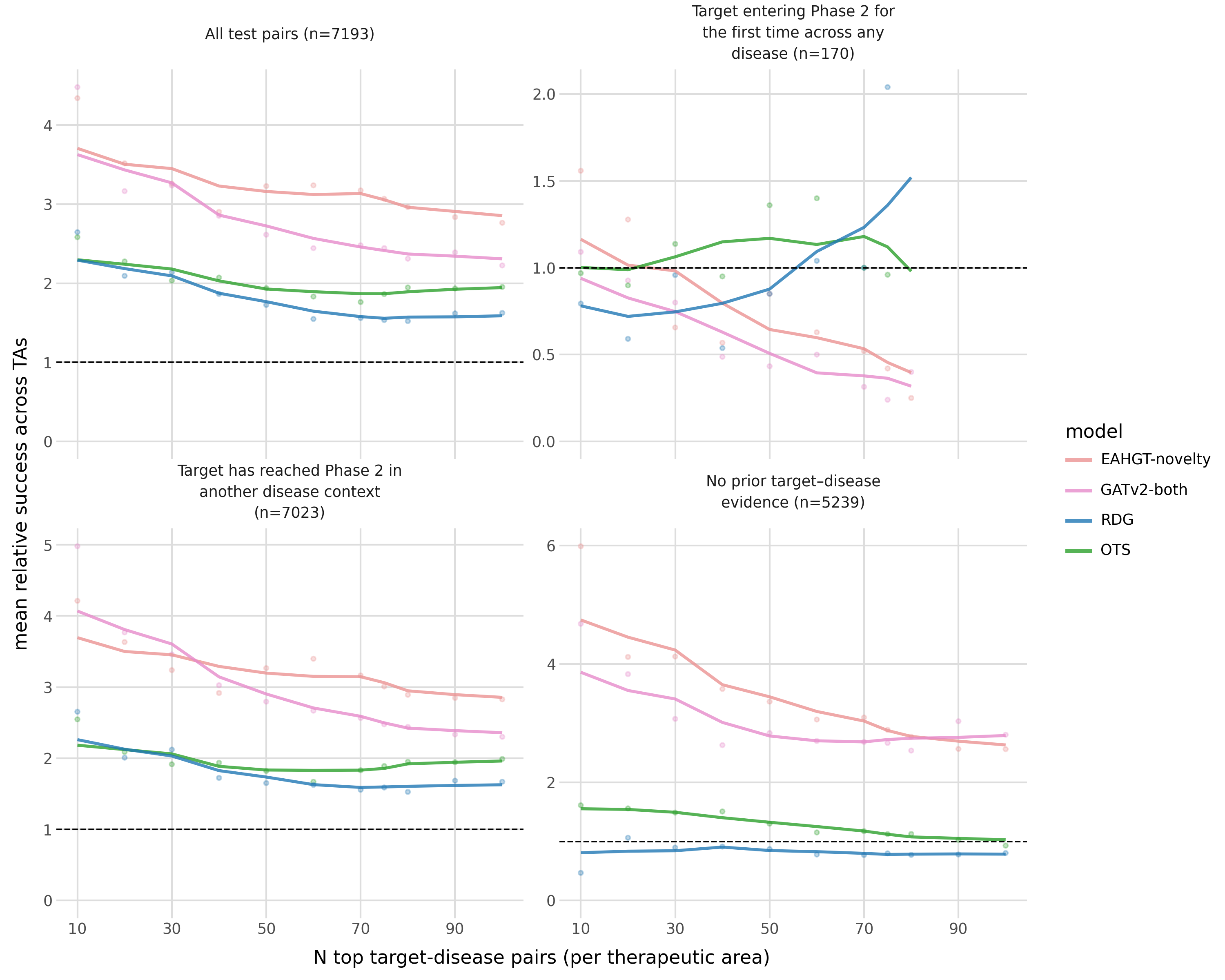}
\caption{Relative success by top-$N$ cutoff within each stratum
(Section~\ref{sec:strata}), for the graph encoders
against the RDG and OTS references. Four panels: all test
pairs ($n=7{,}193$); the two target-clinical-history strata (entering
Phase~II for the first time across any disease, $n=170$; has reached
Phase~II in another disease context, $n=7{,}023$); and the dominant
no-prior-target--disease-evidence stratum ($n=5{,}239$). Panel sub-titles
give the per-stratum pair count.}
\label{fig:rs_by_stratum}
\end{figure}

\paragraph{Performance by evidence stratum}

That advantage is not spread evenly: it varies sharply by stratum, its sign
depending on how much direct evidence a pair carries
(Figure~\ref{fig:rs_by_stratum}; per-stratum values in Appendix
Figure~\ref{fig:rs_by_stratum_heatmap}; strata in Section~\ref{sec:strata}). It is
largest among the dominant \emph{no prior target--disease evidence} group, where
the strongest encoders reach RS@10 $\approx$ 4.7--6.0; five- to sixfold
enrichment over random ranking, recoverable only by propagating over the
intervening biology. The
direct-edge baseline has no pair-specific evidence to read there and falls back on
its target- and disease-level features, which order these pairs no better than
chance: its RS@10 sits \emph{below} the random reference ($\approx$ 0.5).
A paired Wilcoxon over the 13 areas confirms the pattern for the best condition
of each encoder family: the improvement over RDG is significant on the
evidence-free stratum ($p \leq 0.004$) but not on the pooled evaluation set
($p \geq 0.18$). The graph helps where direct evidence is absent and is
indistinguishable from the baseline where it is present.

The target-history axis is underpowered. Only $n = 170$ pairs carry a target
entering Phase~II for the first time in any indication, so a per-area ranking
exhausts the stratum within the reported range: beyond $N \approx 30$ the curves
rank most of it and converge on RS $= 1$ by construction. Within that range the
encoders retain a slight edge at the tightest cutoff, \eahgt{}~($n_{ij}$)
reaches RS@10 $\approx$ 1.2 against $\approx$ 0.8 for RDG but the paired test
over the 13 areas does not reach significance ($p = 0.08$), and we report the
stratum as evidence neither way. The gain seen on the evidence-free stratum does
not appear here, and that asymmetry is disussed in 
Section~\ref{sec:discussion_streetlight}.

\subsection{Ablation: Encoder Family and Edge-Feature Modulation}
\label{sec:ablation}

To separate the contribution of the encoder family from that of the evidential
edge features, Table~\ref{tab:ablation} reports all twelve conditions under one
shared recipe (Appendix~\ref{sec:conditions}), with the RDG and OTS references.

\begin{table}[t]
\caption{Ablation over encoder family and edge-feature subset. Every learned
condition is a five-seed validation-selected rank-fused ensemble under one
shared recipe (identical loss, sampling, seeds, and fusion), varying only the
encoder and the edge-feature subset; RDG and OTS are single seeded references,
regenerated by the pipeline of Czech et al.\ \cite{Czech2024CLINICALFORECASTING}
on the same release and split (Appendix~\ref{app:protocol}). RS@$N$ is the
therapeutic-area mean over 13 areas; AUROC and AP are pooled over all evaluation
pairs. Best value per column in bold. Open Targets release 26.03, evaluation
window w3.}
\label{tab:ablation}
\centering
\footnotesize
\setlength{\tabcolsep}{4pt}
\begin{tabular}{@{}lrrrrr@{}}
\toprule
\textbf{Model} & RS@10 & RS@50 & RS@100 & AUROC & AP \\
\midrule
OTS                                     & 2.58 & 1.94 & 1.95 & 0.55 & 0.12 \\
RDG                                     & 2.65 & 1.73 & 1.62 & 0.57 & 0.13 \\
\midrule
\hgt{}                                  & 3.51 & 2.48 & 2.17 & 0.58 & 0.16 \\
\gatv{}                                 & 4.23 & 2.68 & 2.14 & \textbf{0.62} & 0.16 \\
\rgcn{}                                 & 3.36 & 2.17 & 1.62 & 0.59 & 0.14 \\
\compgcn{}                              & 2.00 & 2.21 & 2.18 & 0.52 & 0.11 \\
\midrule
\eahgt{} ($s_{ij}$)                     & 3.48 & 2.15 & 1.79 & 0.57 & 0.14 \\
\eahgt{} ($n_{ij}$)                     & 4.34 & \textbf{3.23} & \textbf{2.76} & 0.56 & \textbf{0.18} \\
\eahgt{} ($s_{ij},\, n_{ij}$)           & 3.57 & 2.66 & 2.62 & 0.57 & 0.16 \\
\midrule
\gatv{} ($s_{ij}$)                      & 4.07 & 2.33 & 1.91 & 0.59 & 0.15 \\
\gatv{} ($n_{ij}$)                      & 3.56 & 2.30 & 2.06 & 0.60 & 0.16 \\
    \gatv{} ($s_{ij},\, n_{ij}$)            & \textbf{4.48} & 2.61 & 2.22 & 0.61 & 0.17 \\
\bottomrule
\end{tabular}
\end{table}

No single condition stands clear of the rest. On the therapeutic-area mean
\gatv{}~$(s_{ij}, n_{ij})$ leads at $N = 10$ and the novelty-only \eahgt{}
variant at $N = 50$ and $N = 100$, but the across-area intervals of the leading
encoders overlap, and the ordering does not survive re-aggregation: the TA-mean is
an equally-weighted average over 13 high-variance areas, and under a pooled
ranking of all evaluation pairs the spread it shows between \compgcn{} (2.00) and
\gatv{}~$(s_{ij}, n_{ij})$ (4.48) narrows sharply. This is the documented failure
mode of such means \cite{Agarwal2021StatisticalPrecipice,
Dehghani2021BenchmarkLottery}, and we accordingly read the per-area median and the
pooled estimate alongside it (Appendix Figure~\ref{fig:agg_divergence}). What the
three aggregations agree on is coarser: graph propagation clears the tabular
baselines at the top of the ranking, and does so most where the direct
target--disease edge is absent.

The ablation separates the encoders instead by \emph{how} the edge features help
them. \hgt{} gains most from novelty (RS@10 $3.51 \to 4.34$) and little from score
or from both, whereas \gatv{} gains most from both together ($4.23 \to 4.48$) and
is \emph{hurt} by novelty alone ($4.23 \to 3.56$). No subset helps every encoder,
so relation-aware message passing carries most of the signal and the edge features
add a further, architecture-specific lift; these differences sit within the
same overlapping intervals, so we read their direction rather than their size.

\subsection{Decision-Point Subgraphs Support the Hypothesis Rationale}
\label{sec:explainability}

A committee weighing a target--disease pair at the Phase~II gate is assessing a
\emph{hypothesis}, and a rank alone does not tell it what that hypothesis rests
on, through what biology the support travels, or whether that support is
long-established or newly emerging. We therefore decompose predictions into the
evidence that produced them.

We adapt PaGE-Link \cite{Zhang2023PaGELink} to extract the lowest-cost paths
between target and disease under a learned edge mask, capped at the encoder's
four-hop receptive field and fused across the five ensemble seeds, so a recovered
path traverses only edges the model used and only evidence predating the decision.

For IL17F~$\rightarrow$~psoriatic arthritis, decided in 2016, the recovered paths
route through an adjacent \emph{disease}: IL17F's own positive trial in psoriasis,
or its literature co-mention with chronic mucocutaneous candidiasis, reaching
IL17RA, IL12B and PDE4A, which are themselves in trial for or associated with
psoriatic arthritis. The pair is rescued by clinical precedent in a neighbouring
indication, and that route does not exist in a graph carrying only
target--disease edges.

The temporal profile shows that the support on that route was in place before the
decision, and the direct edge was not: the on-path genetic association and
expression edges accumulate through 2009--2014 against a direct target--disease
edge that reaches only 0.2 by the decision year
(Appendix~\ref{app:explainability}). The score is carried by evidence that accrued
on the intermediates, and the increase in that evidence is what the model is
reading.

Both readings, the route and its temporal profile, are available only because the
subgraph is reconstructed at the decision date. A path recovered from the
present-day graph would traverse edges that did not exist when the committee met,
and could not distinguish evidence that motivated a decision from evidence the
decision produced.

%% file: kdd_body/06_discussion.tex

\section{Discussion}

The THBKG models the biomedical evidence landscape as it changes rather than as
it stands, so that a clinical decision can be scored on the evidence its maker
actually held. Three properties of the graph, and one of the encoder we build on
it, bound what the model can be asked to conclude.

\paragraph{The model sits inside the loop it learns from}
\label{sec:discussion_streetlight}

The field searches where the light already falls, and this resource is built from
what it found there. Roughly one protein in three remains functionally
understudied despite potential druggability \cite{Edwards2011TooManyRoads,
Oprea2018UnexploredTherapeutic}. Attention to a
gene is predictable from its chemical and biological properties, and that bias
propagates into publication counts, funding and the drugs developed against
disease-associated genes \cite{Stoeger2018IgnoredGenes}. Effort accumulates
fastest where it has accumulated longest, so the recent surge in genetic
associations has not produced a corresponding rise in \emph{unique} novel targets
\cite{Falaguera2025TemporalDiscovery}. Sponsors decide on the evidence that
exists; those decisions become our labels, and the evidence they were made on
becomes our features.

A model trained on that pairing learns the policy that produced the labels, and
that policy rewards prior precedent. Our results are consistent with this: the
encoders gain clearly on pairs whose target had already reached Phase~II in
another indication, and only marginally, without significance, on
first-time-in-Phase~II targets ($n = 170$; Section~\ref{sec:main_results}), the
stratum a discovery programme would most want, and the one the benchmark is least
able to speak to. Propagation mitigates \emph{direct-evidence} sparsity, transferring
precedent to a pair with no edge to \emph{this} disease, but it cannot manufacture
\emph{research-attention} evidence: the dark targets a discovery programme most
needs are exactly where it has least to draw on. A ranker trained and evaluated on
clinic-reaching targets can therefore sharpen the field's existing focus rather
than broaden it, and the resource we release carries that risk to anyone who
reuses it.

\paragraph{Timestamps record curation, not only discovery}
\label{sec:discussion_timestamps}

An edge is dated by when its evidence entered the record, which is not always
when the underlying science was done. Most of the corpus is literature mined from
Europe PMC and carries a genuine publication date, but curated repositories
supply a submission date instead, and where a resource is integrated in bulk its
contents enter as a block dated to the integration. Falaguera et al.\
\cite{Falaguera2025TemporalDiscovery} show both effects in the evidence we build
on: affected-pathway evidence spikes in 2018 and 2021 with the bulk ingestion of
new datasources rather than with discoveries in those years, and
somatic-mutation evidence declines over the past decade because the Cancer Gene
Census tightened its inclusion criterion. The affected datasources are a minority
of the corpus, but for those a pair can appear evidence-sparse at its decision
point while the supporting studies were already published but not yet curated.

\paragraph{The public hypothesis is not the sponsor's hypothesis}
\label{sec:discussion_proprietary}

The THBKG is assembled from public evidence; the decisions it learns to predict
were not made on public evidence alone. That asymmetry puts the labels and the
features in different information sets: a pair we score as evidence-sparse may
have advanced on internal evidence we cannot see, reaching the model as label
noise. The resource is correspondingly most useful where public evidence is the
binding constraint, in academic target prioritisation and in asking whether
public evidence alone would have supported a decision taken on other grounds.

\paragraph{The encoder compresses the temporal signal it is given}
\label{sec:discussion_compression}

The THBKG retains a dated event stream for every pair, but the encoder does not consume it: each
history is summarised into two hand-designed scalars, a cumulative score and a
novelty score, before message passing begins, which fixes by hand what the
accumulation dynamics of evidence look like. The ablation suggests that choice is
doing real work, since the novelty score alone is the strongest single
edge-feature condition we test (Section~\ref{sec:ablation}). A continuous-time
dynamic graph model \cite{Rossi2020TemporalGraphNetworks} could instead consume
the arrivals directly and learn those dynamics, and the trajectories the THBKG
retains are what would make that possible; that the compressed form already
carries this much signal bounds from below what the uncompressed stream might
carry.

%% file: kdd_body/07_conclusion.tex

\section{Conclusion}

Evaluating a target--disease hypothesis requires traversing the biological network
as it stood when the decision fell due, and no existing biomedical knowledge graph
supports both: those that propagate over multi-hop structure cannot be rewound,
and those that carry timestamps freeze at a single global cutoff or are built from
literature alone. The THBKG closes the gap. Every edge carries the year its
evidence changed, so masking the edges that postdate a pair's
own decision recovers the evidence landscape as that pair's decision-maker saw it,
per instance rather than at one cutoff shared across pairs decided years apart,
while the multi-hop structure needed to reach indirect evidence survives the
masking.

On this graph we define a decision-aligned advancement benchmark and
characterise a suite of reference baselines. The encoders enrich the top of the
ranking over the tabular baselines, and do so most on the evidence-sparse pairs,
where the strongest rank highest a set of pairs that advance at roughly five to
six times the rate expected under random ranking. No single encoder wins outright and
whether the edge features help is architecture-dependent, so we report the suite
as reference points rather than a claimed state of the art. The persistent finding
is not which model wins but where the graph helps: precisely the pairs a
direct-edge model cannot see. Whether it helps on targets reaching Phase~II for
the first time, the regime a discovery programme most wants, is a question this
evaluation set is too thin to settle. A path-based explainer adapted to the decision-time subgraph
recovers what a scalar score withholds: which intermediate biology carries the
signal, by what relation it reaches the disease, and whether the support is
long-settled or newly arrived.

The THBKG enables retrospective validation of drug development decisions: a
hypothesis can be scored on the evidence that stood at a chosen past date, then
checked against what the programme went on to do. Evidence reported later extends
the graph rather than invalidating it, so the set of decisions available to
validate against grows with the record. We demonstrate this at the Phase~II
advancement task, but the query semantics hold for any hypothesis carrying a decision
date, so the substrate extends to other stages of development such as novel target identification and
drug repurposing. We release the THBKG and its pipeline as that substrate.

%% file: kdd_body/A_ethical_use_of_data.tex

\section{Ethical Use of Data}
\label{app:ethics}

The THBKG is derived entirely from publicly available secondary sources
and no new primary data were generated in this study. It contains no
human-subject data, no personally identifiable information, and no
patient-level records; every node is a biological entity (target,
disease, pathway, drug) or a registered clinical-trial identifier, and
every edge is an aggregate association derived from published evidence.
Institutional review and informed consent are therefore not applicable.

Source data are used in accordance with their respective licences. Open
Targets~\cite{Ochoa2021OpenPrioritisation} releases its data under CC0 1.0, and
records that its constituent datasources have agreed to unrestricted use by
Open Targets users; all Open Targets--derived content in the THBKG, including
the clinical-trial and drug-modulation relations built from
\texttt{clinical\_precedence}, is consumed through that release.
ChEMBL~\cite{Gaulton2017The2017} is separately licensed under CC-BY-SA 3.0.
ChEMBL-derived content enters the THBKG only through the CC0-licensed Open
Targets release rather than by direct ingestion, so it is consumed under CC0
terms and the THBKG carries no share-alike obligation; we cite ChEMBL as its
attribution terms request. Clinical-trial records from ClinicalTrials.gov are in the public
domain as a work of the United States federal government.

We note one substantive ethical consideration specific to this resource.
Because the THBKG indexes evidence by the year it entered the public
record, a model trained on it inherits the attention biases of the
literature it is built from: well-studied targets accumulate evidence
faster, and a ranking model rewarded for enrichment may therefore
concentrate on biology that is already well characterised at the expense
of the understudied targets a discovery programme most needs. We treat
this as a property of the resource rather than an artefact of any one
model, discuss it in Section~\ref{sec:discussion_streetlight}, and flag it for
anyone building on the THBKG. We note that we do not measure it: whether
sparser neighbourhoods drive the effect is not something our stratification
resolves.

\section{Data and Code Availability}
\label{app:availability}

The THBKG and its end-to-end construction pipeline, together with the
code for the \eahgt{} encoder and all baseline conditions, are available
at \url{https://github.com/jackysiupuichung/thbkg}
(release tag \texttt{v26.03}) and archived at
\url{https://doi.org/10.5281/zenodo.20795232}. The data are released
under CC-BY-4.0 and the code under the MIT licence. The THBKG is derived
from Open Targets release 26.03. The release includes a Croissant metadata
record (MLCommons Croissant~1.0) describing the graph, the train and
evaluation advancement-label splits, and the label schema, so the benchmark can
be loaded programmatically by standard dataset tooling.

%% file: kdd_body/B_datasheet.tex

\section{Datasheet}
\label{app:datasheet}

We document the THBKG following the \emph{Datasheets for Datasets} schema
\cite{Gebru2021Datasheets}. Counts refer to the Open Targets release~26.03 build;
the released split is the w3 evaluation window (\S\ref{app:protocol}).

\paragraph{Motivation}
The dataset was created to support \emph{decision-aligned} retrospective evaluation of
therapeutic target--disease hypotheses: assessing a hypothesis as of the date its
clinical decision was taken, using only evidence datable before that date.
Existing biomedical knowledge graphs either cannot be rewound to a historical
cutoff or freeze at a single global snapshot, precluding per-instance
decision-aligned evaluation (\S1). The graph and benchmark were assembled by the
authors for the research reported here; no external entity funded a bespoke
collection.

\paragraph{Composition}
Instances are typed nodes and dated, typed edges of a heterogeneous graph. The
graph contains $110{,}396$ nodes across five types (target $19{,}620$; disease
$47{,}030$; molecule $2{,}636$; Gene Ontology term $38{,}495$; Reactome pathway
$2{,}615$) and $\sim\!11.1$M edges across nineteen context relations, sixteen
temporal and three static (Appendix Table~\ref{tab:edge_types_full}). The
prediction targets are $28{,}795$ target--disease \emph{advancement} labels
(whether a pair entering Phase~II advanced beyond it), split temporally into
$21{,}602$ training and $7{,}193$ evaluation pairs; $9.3\%$ of evaluation pairs
are positive. Each temporal edge carries the year its evidence changed and
two continuous features, a cumulative evidential score and a novelty term
(\S\ref{sec:thbkg_construction}); a relation is rewritten in every year its
score differs from the prior year, so its value at any cutoff is the most recent
edge preceding that year. The graph is a derived, aggregated resource: it
contains no personal data, no patient-level records, and no free text; nodes are
public molecular and disease identifiers. It is not a sample of a larger set ---
it is the complete graph induced by the source releases below.

\paragraph{Collection process}
Evidence was drawn from public releases: Open Targets~26.03 (target--disease
association evidence, aggregating genetic, somatic, pathway, animal-model,
expression and literature sources), Reactome, ChEMBL compound--target activity,
and ClinicalTrials.gov trial records for advancement labels and clinical-trial
edges (\S\ref{sec:thbkg_construction}; provenance per relation in Appendix
Appendix Table~\ref{tab:edge_types_full}). Edge dates are the event time reported by the
source (publication year, trial start year, or year of evidence update); the
construction pipeline is released with the dataset. No data were collected from or
about individuals.

\paragraph{Preprocessing / cleaning / labelling}
Evidence records are aggregated per target--disease--datasource and dated;
literature-derived edges are dated by publication year and clinical-trial edges by
resolved trial date (Appendix~\ref{sec:supp_ct_dates}). Advancement labels follow the
protocol of Czech et al.\ \cite{Czech2024CLINICALFORECASTING}: a pair is positive
if it advanced beyond Phase~II, with train/evaluation membership set by clinical
transition year (\S\ref{app:protocol}). Static ontology hierarchies (disease, GO,
Reactome) carry a sentinel timestamp and are never masked. The raw source releases
remain available from their providers; the pipeline reproduces the graph from
them.

\paragraph{Uses}
The dataset supports the advancement benchmark defined here and, more broadly,
any retrospective evaluation of a target--disease hypothesis carrying a decision
date, including novel-target identification and drug repurposing (\S1). Because
the index is keyed to event time, it also supports \emph{as-of} queries at
arbitrary cutoffs and re-evaluation of a hypothesis as its evidence accrues.
Users should not read the advancement label as a measure of biological validity:
it records clinical progression, which is confounded by commercial and
operational factors (\S6), and the graph reflects curation and publication
biases of its public sources (\S6). It should not be used for individual-level
inference; it contains none.

\paragraph{Distribution and maintenance}
The graph, its construction pipeline, the benchmark splits, and the evaluation
harness are released under the licences and identifiers in
Appendix~\ref{app:availability} (data CC-BY-4.0, code MIT; source-data licences
enumerated there). Because the graph is indexed by event time rather than
ingesting a fixed snapshot, later Open Targets releases extend the index rather
than invalidating it: the maintainers refresh the build on each Open Targets
release, version it by the release identifier (the current build is $26.03$), and
retain prior versions at their archived identifiers so that published results
remain reproducible. Issues and contributions are handled through the public
repository (Appendix~\ref{app:availability}).

%% file: kdd_body/C_graph_details.tex

\section{Graph Construction Details}
\label{app:supp}

\begin{table*}[htbp]
\caption{Full edge type specification for the THBKG. All temporal edges
carry a two-dimensional attribute $[s_{ij}, n_{ij}]$ (cumulative score
and Open Targets novelty score) and a snapshot year. Static edges carry
$[1.0, 1.0]$ and no timestamp. The \texttt{advancement} supervision label
(and its reverse mirror) is the only relation excluded from the GNN
context graph; the clinical-trial relations are retained. Outcome edges
carry terminal-date timestamps and ongoing edges carry study-start dates,
so under the temporal mask neither a trial's outcome nor its existence can
enter the graph before it was observable to a decision-maker.}
\label{tab:edge_types_full}
\centering
\resizebox{\linewidth}{!}{%
\begin{tabular}{l l l l r >{\raggedright\arraybackslash}p{6.5cm}}
\toprule
\textbf{Category} & \textbf{Relation} & \textbf{Src} & \textbf{Dst} &
\textbf{Count} & \textbf{Data sources} \\
\midrule
\multirow{6}{*}{\rotatebox[origin=c]{90}{\parbox{2.2cm}{\centering Target--disease\\evidence}}}
& \texttt{literature}            & target & disease & 7{,}269{,}364 & EuropePMC \\
& \texttt{genetic\_association}  & target & disease & 1{,}679{,}195 & GWAS credible sets, EVA, Genomics England, UniProt variants, gene burden, Orphanet, UniProt literature, gene2phenotype, ClinGen \\
& \texttt{animal\_model}         & target & disease & 711{,}880 & IMPC \\
& \texttt{rna\_expression}       & target & disease & 148{,}704 & Expression Atlas \\
& \texttt{somatic\_mutation}     & target & disease & 81{,}193  & Cancer Gene Census, IntOGen, EVA somatic, cancer biomarkers \\
& \texttt{affected\_pathway}     & target & disease & 16{,}570  & Reactome, CRISPR screen, CRISPR, cancer biomarkers \\
\midrule
\multirow{5}{*}{\rotatebox[origin=c]{90}{\parbox{2.2cm}{\centering Clinical\\trials}}}
& \texttt{clinical\_trial\_ongoing}        & target & disease & 76{,}475 & OT \texttt{clinical\_precedence} / CT.gov; dated at \texttt{studyStartDate} \\
& \texttt{clinical\_trial\_positive}       & target & disease & 53{,}101 & \emph{ibid.}; completed, dated at terminal date \\
& \texttt{clinical\_trial\_unknown}        & target & disease & 35{,}572 & \emph{ibid.}; administrative or operational \\
& \texttt{clinical\_trial\_unmet\_efficacy}& target & disease &  4{,}112 & \emph{ibid.}; stop reason contains `negative' \\
& \texttt{clinical\_trial\_adverse\_effects}& target & disease &  2{,}056 & \emph{ibid.}; stop reason contains `safety' \\
\midrule
\multirow{5}{*}{\rotatebox[origin=c]{90}{\parbox{2.5cm}{\centering Pathway,\\functional,\\molecular}}}
& \texttt{interacts\_with}       & target & target   & 659{,}443 & IntAct; dated at earliest reporting publication \\
& \texttt{has\_function\_in}     & target & go       & 129{,}649 & Gene Ontology annotations; dated at earliest supporting publication (annotations without a resolvable reference are not admitted) \\
& \texttt{modulated\_by}         & molecule & target &  75{,}477 & OT \texttt{clinical\_precedence} / CT.gov; the drug--target arm of the same trial records, so clinical development rather than binding affinity \\
& \texttt{involved\_in}          & target & reactome &  4{,}994 & Reactome; Open Targets evidence date \\
& \texttt{associated\_with}      & disease & reactome &  969 & Reactome; Open Targets evidence date \\
\midrule
Label
& \texttt{advancement}           & target & disease &  28{,}795 & CT.gov trials mapped to targets via ChEMBL (\S\ref{sec:advancement_labels}); supervision label, not a context relation --- excluded from message passing ($21{,}602$ train $+$ $7{,}193$ evaluation) \\
\midrule
\multirow{3}{*}{\rotatebox[origin=c]{90}{\parbox{1.5cm}{\centering Static\\ontology}}}
& \texttt{is\_subtype\_of}       & disease & disease     & 64{,}093 & EFO / Disease Ontology \\
& \texttt{is\_subtype\_of}       & go & go               & 58{,}384 & Gene Ontology (\texttt{is\_a}) \\
& \texttt{is\_subpathway\_of}    & reactome & reactome   &  2{,}633 & Reactome \\
\bottomrule
\end{tabular}%
}
\end{table*}

\subsection{Clinical-Trial Edge Date Resolution}
\label{sec:supp_ct_dates}

Each resolvable trial emits one outcome edge (positive, unmet efficacy,
adverse effects, or unknown/operational) at its terminal date, taken as the
first available of \texttt{resultsFirstPostDate},
\texttt{lastUpdatePostDate}, the actual completion date, then the estimated
completion date, and clipped to be no earlier than the study start date.
Outcome categories are mapped from the ChEMBL \texttt{clinicalStatus} and
\texttt{studyStopReasonCategories} fields following
\cite{Narganes-Carlon2024GATher:Links}. The ongoing edge is dated at the
\texttt{studyStartDate}.

\begin{table}[htbp]
\centering
\caption{Node types in the THBKG with canonical identifiers, entity
counts, and feature dimensionality. Counts are nodes carrying at least one
edge; identifiers present in a source release but never realised as an edge
are not admitted to the graph. Feature construction per type is
described in Appendix Section~\ref{sec:supp_features}.}
\label{tab:ch4-node-types}
\begin{tabular}{@{}llrr@{}}
\toprule
\textbf{Node type} & \textbf{Identifier} & \textbf{Count} &
\textbf{Feat.\ dim} \\
\midrule
Target   & Ensembl gene ID & 19{,}620 & 56      \\
Disease  & EFO / MONDO ID  & 47{,}030 & 256     \\
GO term  & GO term ID      & 38{,}495 & 64      \\
Molecule & ChEMBL ID       &  2{,}636 & 1{,}024 \\
Reactome & Reactome ID     &  2{,}615 & 64      \\
\bottomrule
\end{tabular}
\end{table}

\subsection{Node Feature Construction}
\label{sec:supp_features}

Each node type is initialised with a feature vector derived from its
biological or chemical properties, following
\cite{Narganes-Carlon2024GATher:Links}.
\textbf{Targets} ($d = 56$) combine gene-expression specificity and
evolutionary constraint: transcript abundance across 81 cell types,
quantified as normalised Transcripts Per Million from single-cell RNA
sequencing and summarised by the Jensen--Shannon Specificity score, plus
genetic-constraint features (Haploinsufficiency and Triploinsufficiency
scores, the pLI/pLOEUF loss-of-function-tolerance score, and a Common
Essential flag).
\textbf{Diseases} ($d = 256$) are sentence embeddings of EFO names and
descriptions \cite{Ochoa2021OpenPrioritisation} from a pre-trained
biomedical language model, encoding semantic proximity so the encoder can
generalise across related conditions when graph connectivity is sparse.
\textbf{GO terms and Reactome pathways} ($d = 64$ each) are graph
auto-encoder embeddings of their ontology hierarchies (\texttt{is\_a} /
\texttt{part\_of} for GO; \texttt{is\_subpathway\_of} for Reactome),
preserving structural proximity within each ontology.
\textbf{Molecules} ($d = 1{,}024$) are Morgan (circular) fingerprints from
canonical SMILES at radius 2 with a 1{,}024-bit vector, encoding local
chemical substructure.

%% file: kdd_body/D_models.tex

\section{Model Details}
\label{app:models}

\subsection{Edge-Aware Attention (\eahgt{})}
\label{app:eahgt}

\hgt{} \cite{Hu2020HeterogeneousTransformer} weights two edges of the same
relation type identically, regardless of the evidential score $s_{ij}$ or novelty
$n_{ij}$ they carry. \eahgt{} modulates \hgt{}'s multiplicative attention by these
edge features:
\begin{equation}
    \alpha_{ts} = \operatorname{softmax}_{s}\!\left(
        \frac{
            \left(\mathbf{Q}^{\tau(t)} \mathbf{h}_t\right)^\top
            \mathbf{W}^{\phi(e)}_{\mathrm{ATT}}\,
            \mathbf{K}^{\tau(s)} \mathbf{h}_s
        }{\sqrt{d}}
        \;\cdot\;
        \left(\mathbf{w}^{\phi(e)}_e\right)^\top \mathbf{f}_{ts}
    \right),
    \label{eq:eahgt-attn}
\end{equation}
where $\mathbf{w}^{\phi(e)}_e \in \mathbb{R}^2$ is a learned per-relation
projection of the edge feature onto a scalar gate. The message and aggregation
steps are unchanged from \hgt{}, and the modification adds two parameters per
relation type. Each relation is mirrored by a type-distinct reverse relation
carrying the same timestamp and features, permitting bidirectional message
passing.

Attention is modulated by the edge features but not by the event times
themselves. Per-instance masking (Equation~\eqref{eq:asof}) already clips the
subgraph to the decision point, and conditioning attention on the years of
masked-out edges would reintroduce the post-decision information the masking
removes.

\begin{table}[htbp]
\centering
\caption{Hyperparameters for the primary \eahgt{} ($s_{ij},\, n_{ij}$)
configuration.}
\label{tab:hyperparams}
\begin{tabular}{@{}ll@{}}
\toprule
\textbf{Parameter} & \textbf{Value} \\
\midrule
\multicolumn{2}{@{}l}{\textit{Encoder}} \\
Hidden / output dimension & 128 \\
Attention heads & 2 \\
Number of HGT layers & 2 \\
Dropout (encoder) & 0.2 \\
Edge feature dimension & 2 ($s_{ij}$, $n_{ij}$) \\
Neighbourhood sample (L1 / L2) & 20 / 10 \\
\midrule
\multicolumn{2}{@{}l}{\textit{Decoder}} \\
MLP layers & 256 $\rightarrow$ 128 $\rightarrow$ 64 $\rightarrow$ 1 \\
Dropout (decoder) & 0.1 \\
\midrule
\multicolumn{2}{@{}l}{\textit{Training}} \\
Loss & LambdaLoss / LambdaRank, TA-pooled \\
 & ($k = 100$, $\sigma = 1.43$) \\
Graph orientation & undirected (reverse edges added) \\
Optimiser & AdamW \\
Learning rate & $4.35 \times 10^{-4}$ \\
Weight decay & $1.83 \times 10^{-3}$ \\
LR scheduler & CosineAnnealingLR ($\eta_{\min} = 10^{-6}$) \\
Batch size & 512 \\
Gradient clipping & max\_norm $= 1.0$ \\
Checkpoint selection & max val per-TA median NDCG@50 \\
Ensemble & 5 seeds (1, 7, 42, 123, 2024), \\
 & percentile-rank fusion \\
\bottomrule
\end{tabular}
\end{table}


\subsection{Experimental Conditions}
\label{sec:conditions}

To isolate the contributions of the graph encoder and of the continuous
edge features independently, we define the following conditions under a
shared evaluation protocol on the same graph and split. Four
baseline encoders (\hgt{}, \gatv{}, \rgcn{}, \compgcn{}) span the main
families of heterogeneous and multi-relational message passing without
edge-feature conditioning; the two attention-based backbones (\hgt{} and
\gatv{}) then each take the edge features in three subsets, with the
cumulative-score and novelty-only variants isolating each feature's
individual contribution and the third using both jointly, giving a
$2 \times 3$ ablation. All twelve conditions are evaluated on the 26.03
build and the w3 split under one shared ensembling recipe
(Section~\ref{sec:ablation}).

\begin{table}[htbp]
\caption{Summary of experimental conditions. $s_{ij}$: cumulative
evidential score. $n_{ij}$: Open Targets novelty score. ``Edge
features'' denotes whether the continuous per-edge attributes enter
message passing. Upper block: baseline encoders without edge features.
Lower blocks: the two attention backbones with each edge-feature
subset. The claim each condition tests is discussed in the text and
analysed in Section~\ref{sec:ablation}.}
\label{tab:conditions}
\centering
\small
\begin{tabular*}{\columnwidth}{@{}l@{\extracolsep{\fill}}l@{}}
\toprule
\textbf{Model} & \textbf{Edge features} \\
\colrule
\hgt{}                                    & None \\
\gatv{}                                   & None \\
\rgcn{}                                   & None \\
\compgcn{}                                & None \\
\colrule
\eahgt{} (cumulative score only)          & $s_{ij}$ \\
\eahgt{} (novelty only)                   & $n_{ij}$ \\
\eahgt{} (cumulative $+$ novelty)         & $s_{ij},\, n_{ij}$ \\
\colrule
\gatv{} (cumulative score only)           & $s_{ij}$ \\
\gatv{} (novelty only)                    & $n_{ij}$ \\
\gatv{} (cumulative $+$ novelty)          & $s_{ij},\, n_{ij}$ \\
\botrule
\end{tabular*}
\end{table}

\noindent Comparing each backbone's edge-aware variants against the same
backbone without edge features isolates the contribution of the
edge-feature modulation; contrasting the four
edge-feature-free encoders among themselves compares encoder families at
matched inputs; and running the same three subsets on two backbones
separates a feature's contribution from the architecture that consumes
it, measuring the combined effect of the encoder and the THBKG's evidential edge
attributes.

\paragraph{Hyperparameters}

Encoder capacity and learning hyperparameters were selected by grid search
and Optuna, chosen by NDCG@50 on the validation split under the
decision-aligned regime (full configuration in Appendix
Table~\ref{tab:hyperparams}); neighbourhood sampling draws the most recent
20 then 10 edges per type at layers 1 and 2. All experiments ran on the QMUL
Apocrita HPC cluster on a single NVIDIA A100 GPU.


%% file: kdd_body/E_protocol.tex

\section{Training and Evaluation Protocol}
\label{app:protocol}

\paragraph{Prediction head and objective}
After $L = 2$ message-passing layers, the target and disease embeddings
($d = 128$; Appendix Table~\ref{tab:hyperparams}) are concatenated and passed
through a three-layer MLP decoder (ReLU, dropout $0.1$) to a scalar
ranking logit. We train with the LambdaLoss formulation of LambdaRank
\cite{Wang2018TheOptimization, Pobrotyn2020ContextAwareLearning} (slope
$\sigma = 1.43$, NDCG truncated at $k = 100$), which weights pairwise
losses by the NDCG change of each swap and so aligns training directly
with the top-of-list RS@$N$ metric. All conditions rank pairs as one
therapeutic-area--pooled list.

\paragraph{Multi-seed validation-selected ensemble}
The temporal distribution shift between train and test weakens the
early-stopping signal, since no validation metric reliably tracks the test
peak \cite{Gulrajani2021DomainBed}, so a single run is an unstable
estimate of RS@$N$. We therefore ensemble five independently-seeded runs
(seeds 1, 7, 42, 123, 2024), a remedy known to improve and stabilise model
selection under distribution shift
\cite{Arpit2022EnsembleAverages, Seligmann2023BeyondDeepEnsembles}. For
each run we take the checkpoint maximising the validation per-area median
NDCG@50 (a checkpoint a deployer could actually select, never the test
peak) and fuse runs by averaging per-pair \emph{percentile ranks}: rank
fusion is scale-free, whereas averaging the differently-scaled LambdaRank
logits would let one run dominate. The rule uses no test-set information
and is fully deployable.

\paragraph{Dataset splits}

The labelled pairs (Section~\ref{sec:advancement_labels}) are partitioned
by year of first Phase~II entry, following Czech et al.\
\cite{Czech2024CLINICALFORECASTING}: the \textbf{training set} is pairs
entering Phase~II in 1995--2015, the \textbf{evaluation set} pairs
entering in 2016--2021, so that every evaluation pair has the follow-up
needed to observe advancement. On our Open Targets 26.03 build this yields
$21{,}602$ training pairs, of which 22.76\% (4{,}917) advanced, and
$N = 7{,}193$ evaluation pairs, of which 9.29\% (668) did.

The positive rate is thus far lower in evaluation than in training. The gap
follows from the label definition rather than from the split: a pair counts as
advancing only if a Phase~III trial is registered within three years of its
Phase~II entry (Section~\ref{sec:advancement_labels}), so pairs entering late in
the evaluation window have less of that window observed and fall to the negative
class. The evaluation set is correspondingly the harder and sparser half, and the
shift it induces is why no validation metric reliably tracks the test peak and why
every condition is ensembled over five seeds rather than early-stopped on a single
run. Table~\ref{tab:traineval_ta} resolves both splits by therapeutic area; the
per-area evaluation rate ranges from 4.30\% to 27.87\%, which is why RS@$N$ is
reported as a therapeutic-area mean rather than pooled
(Section~\ref{sec:metrics}).

\begin{table}[htbp]
\caption{Advancement labels by therapeutic area, Open Targets 26.03,
evaluation window w3. Counting is \emph{multi-membership}: a pair is counted in
every therapeutic area its disease maps to, matching the RS-by-area computation,
so rows overlap and do not sum to the split totals --- the \emph{all} row gives
the overlapping totals across primary areas, and the de-duplicated totals are
$21{,}602$ training pairs (4{,}917 positive, 22.76\%) and $7{,}193$ evaluation
pairs (668 positive, 9.29\%). The 13 primary areas are shown; roughly 15\% of
pairs map only to non-primary areas and are omitted.}
\label{tab:traineval_ta}
\centering
\footnotesize
\setlength{\tabcolsep}{3.5pt}
\begin{tabular}{@{}lrrrrr@{}}
\toprule
& \multicolumn{2}{c}{\textbf{Train}} & \multicolumn{3}{c}{\textbf{Evaluation}} \\
\cmidrule(lr){2-3}\cmidrule(lr){4-6}
\textbf{Therapeutic area} & Pairs & \% pos & Pairs & Pos. & \% pos \\
\midrule
\emph{all} (overlapping)             & 42{,}524 & 22.49 & 14{,}227 & 1{,}323 & 9.30 \\
\midrule
Cancer or benign tumour              & 10{,}511 & 20.87 & 2{,}535 & 192 & 7.57 \\
Genetic, familial or congenital      &  3{,}964 & 22.75 & 1{,}755 & 222 & 12.65 \\
Nervous system                       &  3{,}886 & 21.02 & 1{,}312 &  75 & 5.72 \\
Immune system                        &  2{,}802 & 21.81 &    987 & 127 & 12.87 \\
Haematologic                         &  2{,}735 & 20.22 &    949 & 129 & 13.59 \\
Musculoskeletal or connective tissue &  2{,}397 & 19.69 &    902 & 113 & 12.53 \\
Gastrointestinal                     &  3{,}123 & 23.98 &    790 &  34 & 4.30 \\
Endocrine system                     &  3{,}182 & 22.44 &    713 &  44 & 6.17 \\
Cardiovascular                       &  1{,}682 & 18.85 &    612 &  36 & 5.88 \\
Reproductive system or breast        &  2{,}083 & 24.29 &    584 & 153 & 26.20 \\
Urinary system                       &  1{,}356 & 18.58 &    574 & 160 & 27.87 \\
Integumentary system                 &  1{,}271 & 23.13 &    571 &  98 & 17.16 \\
Respiratory or thoracic              &  1{,}368 & 26.75 &    520 &  44 & 8.46 \\
\bottomrule
\end{tabular}
\end{table}

\paragraph{Provenance of the tabular references}

RDG and OTS are regenerated for this benchmark rather than transcribed from
their published values. Both are produced by the pipeline of Czech et al.\
\cite{Czech2024CLINICALFORECASTING}, run over the Open Targets 26.03 release
and the evaluation split above with only the adaptations that release's schema
requires, leaving the feature construction, the ridge specification and the
fitting procedure unchanged. Every condition in Table~\ref{tab:ablation} is
therefore fitted and scored on the same pairs. Because the underlying release
differs from the one available to Czech et al., the values reported here are not
directly comparable to their published figures: the difference is one of data
version, not of model or implementation.

\paragraph{Therapeutic area stratification}

Results are reported as equally-weighted averages across a subset of
therapeutic areas, following the selection criteria of Czech et al.\
\cite{Czech2024CLINICALFORECASTING}: a therapeutic area is retained if it contains at least
100 target--disease pairs with direct target--disease-specific evidence
of any kind within the evaluation set, following Czech et al.'s two filters. A
\emph{name} filter first drops six areas that are not disease systems relevant to
target prioritisation --- \emph{animal disease}, \emph{biological process},
\emph{infectious disease}, \emph{injury, poisoning or other complication},
\emph{medical procedure}, and \emph{pregnancy or perinatal disease} --- by name,
irrespective of their evidence count (\emph{infectious disease} clears the count
threshold but is excluded here). A \emph{count} filter then drops five areas with
fewer than 100 evidence-bearing pairs: \emph{psychiatric disorder} (96),
\emph{pancreas disease} (89), \emph{phenotype} (76), \emph{nutritional or
metabolic disease} (63), and \emph{disorder of visual system} (47). This yields 13
retained areas, matching the set used in Czech et al.'s primary performance
figures.

%% file: kdd_body/F_results_figures.tex

\section{Supplementary Results}
\label{app:figures}

\begin{figure}[t]
\centering
\includegraphics[width=\columnwidth]{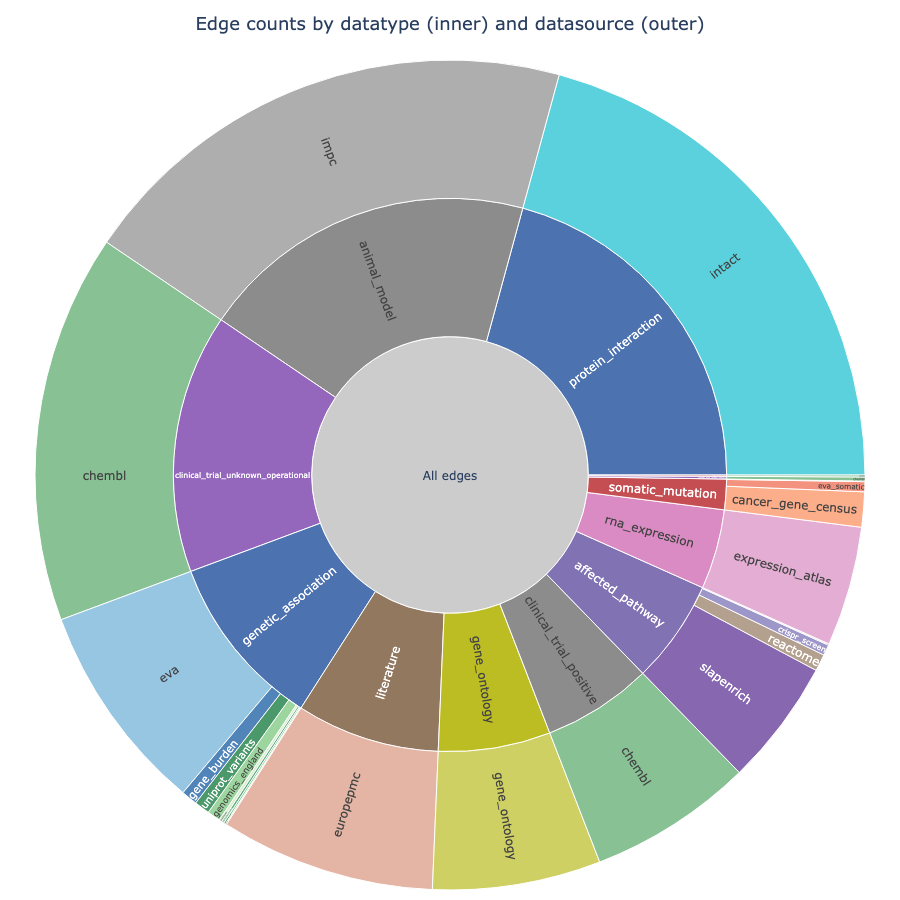}
\caption{Static edge distribution of the THBKG across evidence datatypes and
datasources. The inner ring partitions edges by the seven high-level Open Targets
evidence categories; the outer ring resolves individual datasources. Arc area is
proportional to edge count.}
\label{fig:thbkg_sunburst}
\end{figure}

\begin{figure*}[p]
\centering
\includegraphics[width=0.60\textwidth]{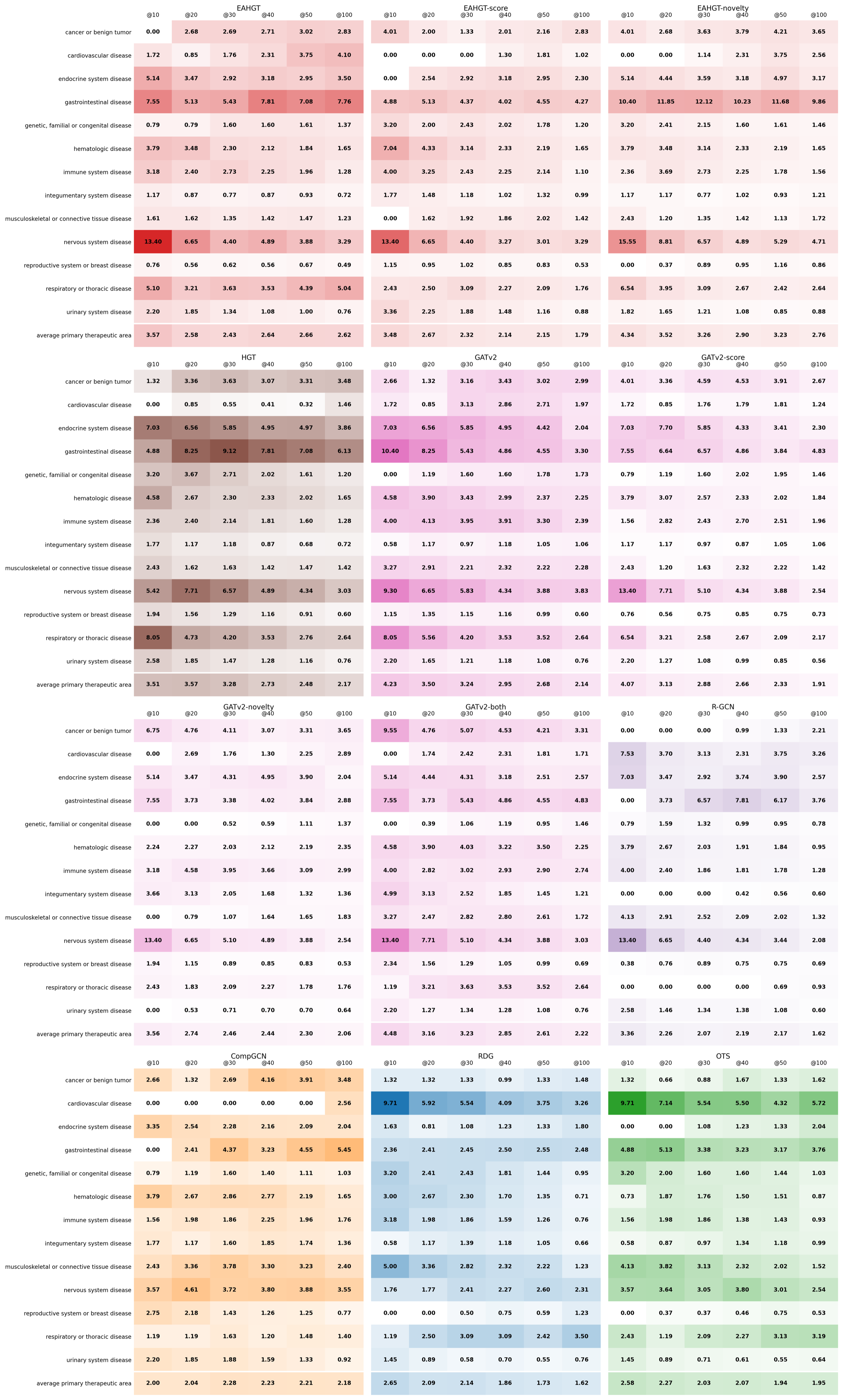}
\caption{Relative success by therapeutic area and top-$N$ cutoff for the
graph encoders and the RDG and OTS references (Open Targets 26.03,
evaluation window w3). One panel per model; within each panel, rows are the
13 primary therapeutic areas and their equally-weighted average, columns are
the cutoff ($N \in \{10, 20, 30, 40, 50, 100\}$), and each cell is the RS
value, shaded by model hue and RS intensity. This gives the full per-area
numbers underlying the per-therapeutic-area analysis
(Section~\ref{sec:main_results}).}
\label{fig:rs_by_ta_heatmap}
\end{figure*}

\begin{figure*}[htbp]
\centering
\includegraphics[width=0.60\textwidth]{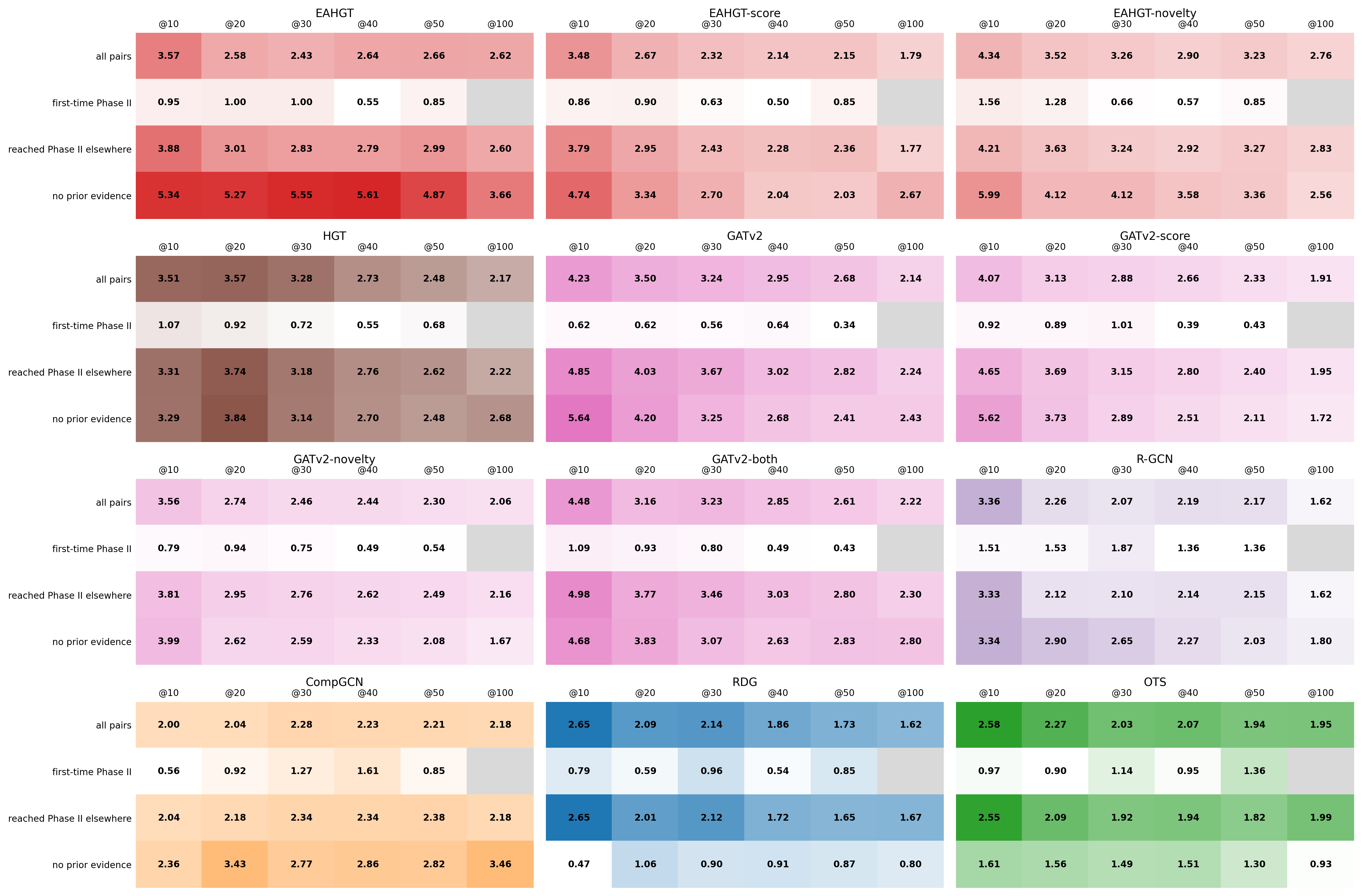}
\caption{Relative success by evidence/clinical-history stratum and top-$N$
cutoff for the graph encoders and the RDG and OTS references (Open Targets
26.03, evaluation window w3). One panel per model; within each panel, rows
are the pooled set and the evidence-availability and target-clinical-history
strata (Section~\ref{sec:strata}), columns are the cutoff
($N \in \{10, 20, 30, 40, 50, 100\}$), and each cell is the TA-mean RS,
shaded by model hue and RS intensity; a grey cell is a stratum with too few
pairs to rank at that cutoff. This gives the per-stratum numbers
underlying the by-stratum analysis (Section~\ref{sec:main_results}).}
\label{fig:rs_by_stratum_heatmap}
\end{figure*}

\begin{figure}[htbp]
\centering
\includegraphics[width=\columnwidth]{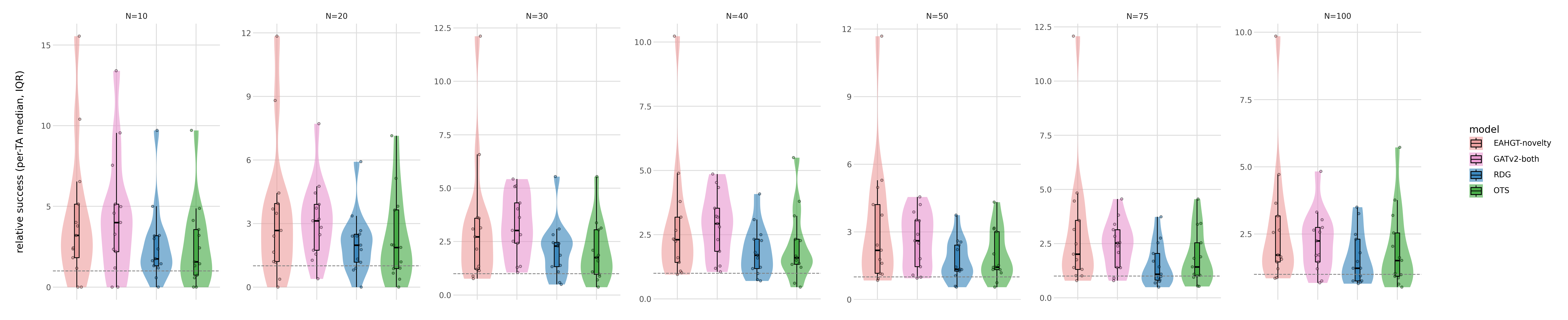}
\caption{Per-therapeutic-area distribution of RS@$N$ at
$N \in \{10, 20, 30, 40, 50, 75, 100\}$ for the headline model set against the
RDG and OTS references (Open Targets 26.03, evaluation window w3): the best
condition per encoder family. Each box shows the median and IQR across the 13
therapeutic areas and each dot is one area. The per-area intervals overlap
across the suite (Section~\ref{sec:ablation}).}
\label{fig:rs_distributions_ta}
\end{figure}

\begin{figure*}[htbp]
\centering
\includegraphics[width=0.85\textwidth]{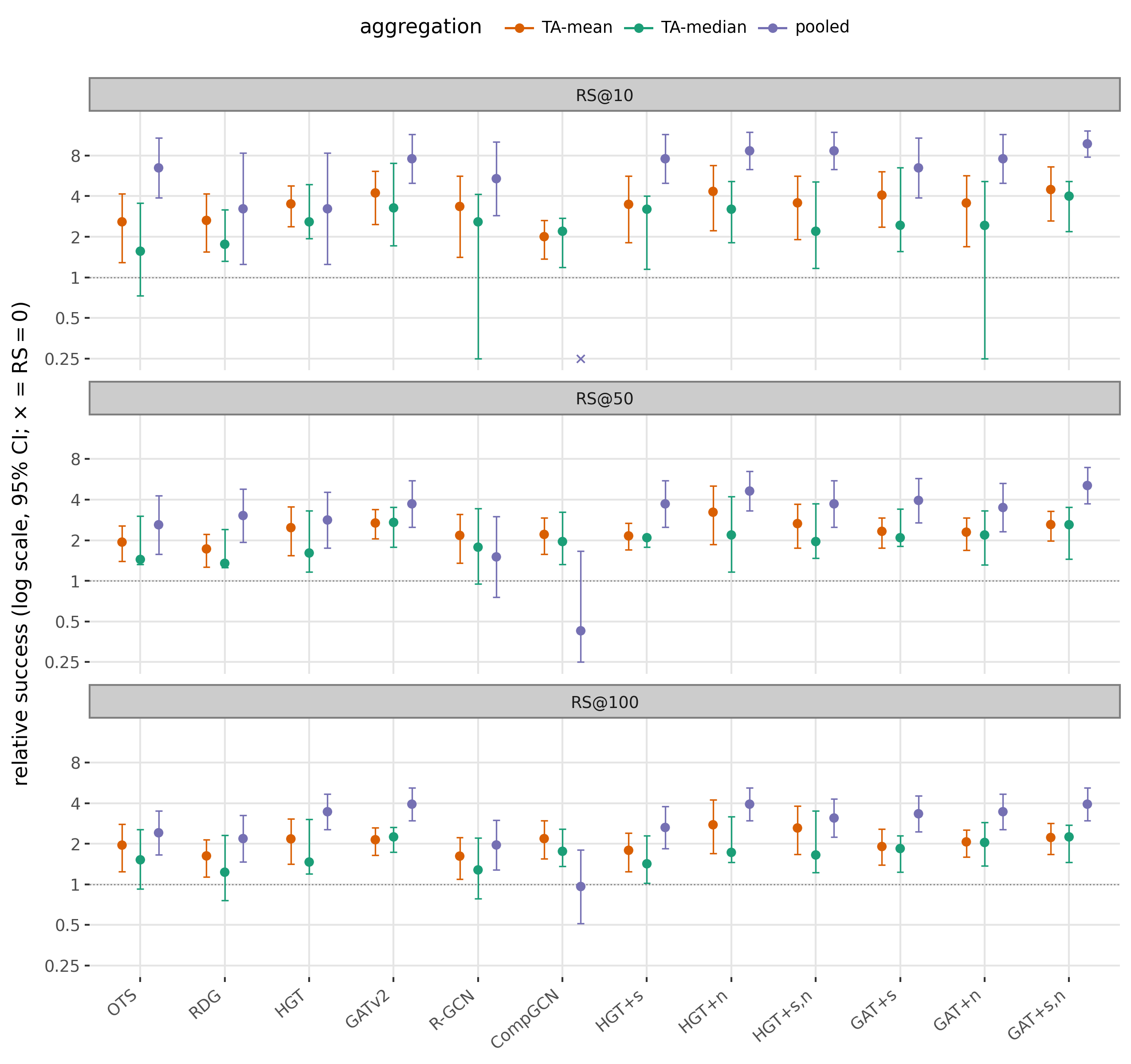}
\caption{Aggregation-method divergence for the ablation conditions of
Table~\ref{tab:ablation}. For each model and cutoff
($N \in \{10, 50, 100\}$), relative success is shown under three
aggregations over the 13 primary therapeutic areas: the equally-weighted
TA-\emph{mean} (macro; spike-prone), the per-TA \emph{median}
(outlier-robust), and the \emph{pooled} single-global-list metric
(micro). Error bars are 95\% intervals; an $\times$ marks conditions with no
advancer in the pooled top-$N$ (RS $=0$), and the $y$-axis is logarithmic. The
three aggregations do not agree on an ordering of the conditions: several are
favoured by the TA-mean alone, the signature of a few high-variance areas
dominating an equally-weighted average over a small number of groups
\cite{Agarwal2021StatisticalPrecipice, Dehghani2021BenchmarkLottery}. This is
why we read the three together and claim only what they share
(Section~\ref{sec:ablation}).}
\label{fig:agg_divergence}
\end{figure*}


\subsection{Aggregate Ranking and Classification Behaviour}
\label{app:aggregate}


\paragraph{Global classification metrics}

The pooled classification result reported in
Section~\ref{sec:main_results} is dominated by the 72.8\% of pairs carrying
no target--disease evidence edge, on which near-chance separability is
expected. The per-therapeutic-area distribution
(Figure~\ref{fig:classification_by_ta}) separates the conditions no better:
the per-area intervals overlap across the suite, and we draw no conclusion
from the ordering.

\paragraph{Per-therapeutic-area analysis}

The encoders rank ahead of RDG in most therapeutic areas but unevenly
(Figure~\ref{fig:rs_distributions_ta}; per-area values in
Figure~\ref{fig:rs_by_ta_heatmap}), and the areas that favour the baseline
differ by encoder. Where a paired test over 13 areas does not reach
significance it reflects limited power rather than an established null. Read alongside the pooled estimate and the
per-area median (Figure~\ref{fig:agg_divergence}), the aggregations agree
only on the coarser statement made in Section~\ref{sec:ablation}: graph
propagation clears the tabular baselines at the top of the ranking, and no one
encoder is separable from the others.

%% file: kdd_body/G_case_studies.tex


\section{Explanation Path Case Studies}
\label{app:explainability}


The path-decomposition procedure of Section~\ref{sec:explainability} is
instantiated below on a prediction whose direct target--disease edge is
near-empty at the decision point.

\begin{figure}[t]
\centering
\includegraphics[width=\columnwidth]{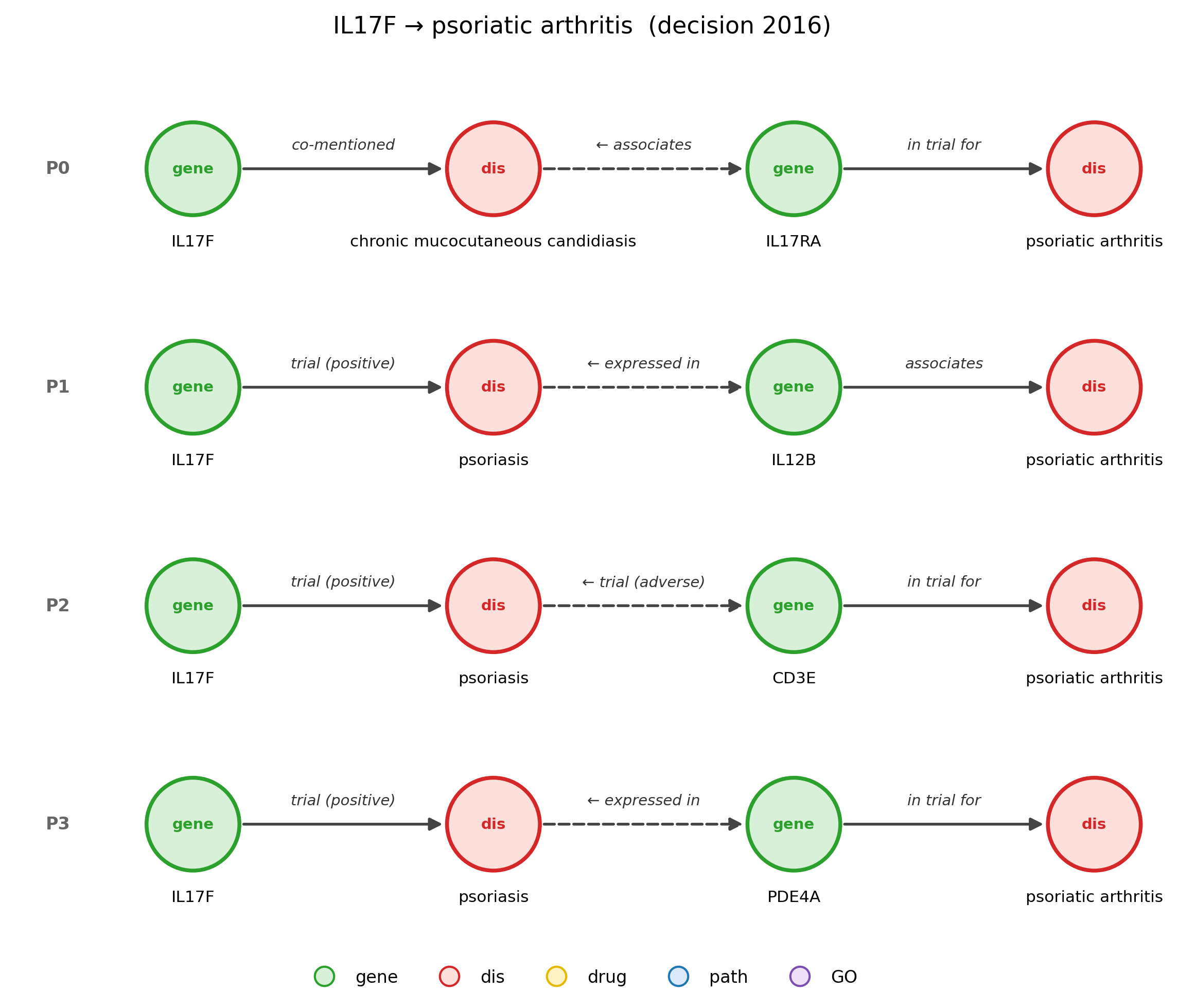}
\caption{Recovered explanation paths for IL17F $\rightarrow$ psoriatic arthritis
(2016). The near-tautological one-hop direct edge is omitted; the paths shown
are the multi-hop routes between target and disease. Node colour denotes
entity type, edge labels give the schema relation, and a dashed arrow marks
an edge traversed backwards. Per-edge evidence strength over time is given
in Figure~\ref{fig:casestudy_temporal}.}
\label{fig:casestudy_paths}
\end{figure}

\begin{figure}[t]
\centering
\includegraphics[width=\columnwidth]{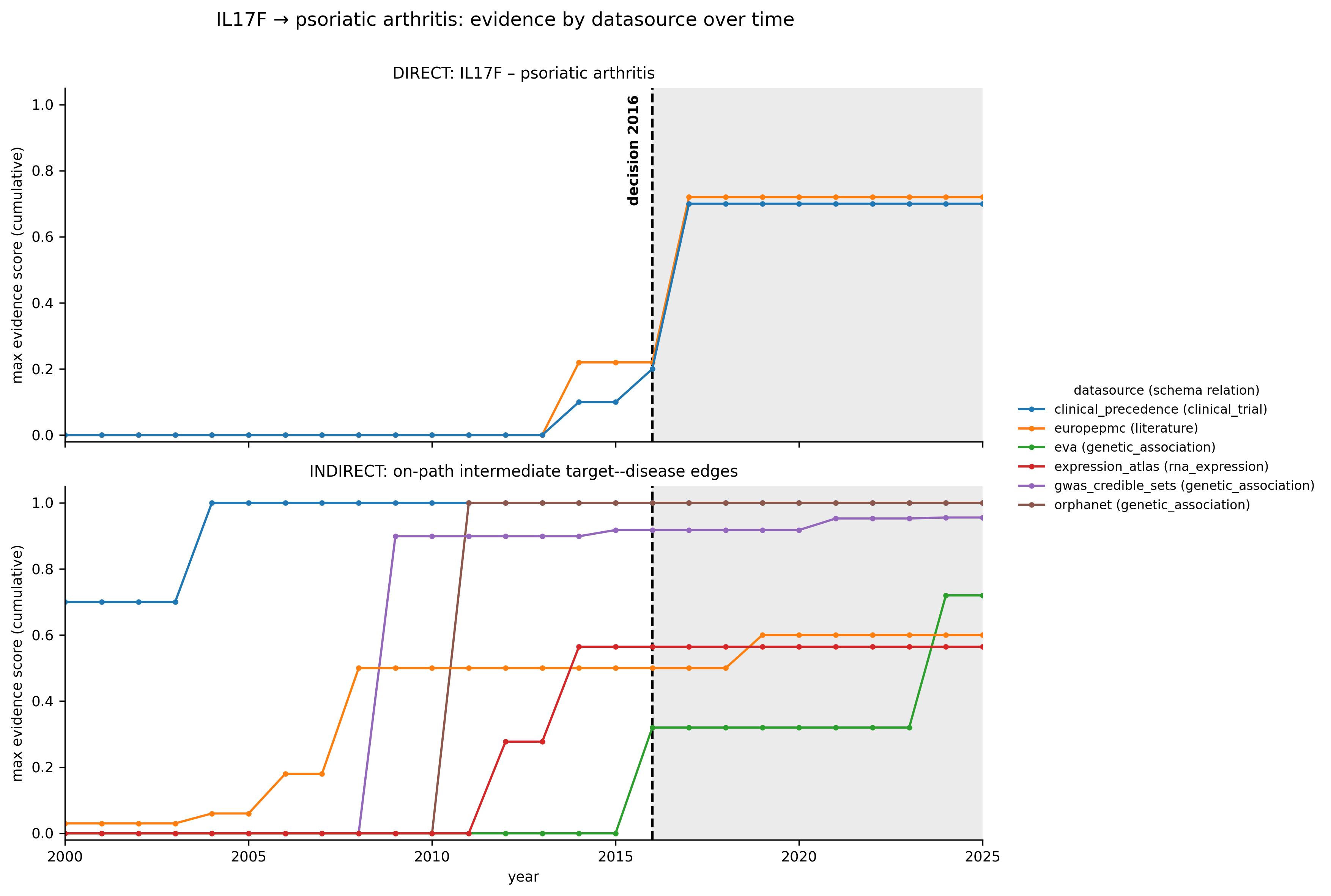}
\caption{Evidence available over time for the prediction of
Figure~\ref{fig:casestudy_paths}, IL17F $\rightarrow$ PsA. The panel plots the
direct target--disease edge (top) and the on-path neighbour evidence the
explanation routes through (bottom), by datasource; the dashed line marks
the decision year.}
\label{fig:casestudy_temporal}
\end{figure}

Applied to predictions whose direct target--disease edge is near-empty at
the decision point, the decomposition returns a recurring structure: the
score is carried by on-path evidence that \emph{leads} the direct edge in
time, so the model is extrapolating from a mature neighbourhood onto an
emerging pair rather than reading a signal that has already arrived
(Figures~\ref{fig:casestudy_paths}, \ref{fig:casestudy_temporal}). For
\textit{IL17F}~$\rightarrow$~\emph{psoriatic arthritis} (PsA), decided in
2016, the \emph{route} is an adjacent disease: paths reach PsA
through a disease IL17F already touches (psoriasis or chronic mucocutaneous
candidiasis) and a target on that disease itself tied to PsA (IL17RA or
IL12B). Its \emph{temporal profile} is recently-grown support, $397$
pre-decision on-path records across genetics, expression, literature and
clinical precedence, much of it accruing through the early 2010s just ahead
of the cut, against a near-flat direct edge (peak score $0.22$). The
decomposition thus recovers not only the route around the empty direct edge but
the vintage of the support on it: here, evidence accumulated shortly before the
decision rather than long-settled precedent.

